\documentclass[letterpaper]{article} 
\usepackage{aaai23}  

\usepackage{times}  
\usepackage{helvet}  
\usepackage{courier}  
\usepackage[hyphens]{url}  
\usepackage{graphicx} 
\usepackage{natbib}  
\usepackage{caption} 
\usepackage{algorithm}
\usepackage{algorithmic}

\usepackage{tikz}
\usepackage{comment}
\usepackage{amsmath,amssymb} 
\usepackage{color}
\usepackage{array}
\usepackage{booktabs}
\usepackage{subcaption}
\usepackage{multirow}
\newcommand{\argmax}{\operatornamewithlimits{argmax}}

\usepackage{newfloat}
\usepackage{listings}
\DeclareCaptionStyle{ruled}{labelfont=normalfont,labelsep=colon,strut=off} 
\floatstyle{ruled}
\newfloat{listing}{tb}{lst}{}
\floatname{listing}{Listing}
\title{Unsupervised Prompt Learning for Vision-Language Models}
\author{
    Tony Huang\textsuperscript{\rm 1}\equalcontrib,
    Jack Chu\textsuperscript{\rm 1}\equalcontrib,
    Fangyun Wei\textsuperscript{\rm 2}\equalcontrib\thanks{Corresponding author.}
}
\affiliations{
    \textsuperscript{\rm 1}Peking University ~~\textsuperscript{\rm 2}Microsoft Research Asia\\
    tonyhuang$\_$pku@outlook.com ~~~~~ chuxicloud@icloud.com ~~~~~ fawe@microsoft.com
}

\usepackage{bibentry}

\begin{document}

\maketitle

\begin{abstract}
Contrastive vision-language models like CLIP have shown great progress in transfer learning. In the inference stage, the proper text description, also known as prompt, needs to be carefully designed to correctly classify the given images. In order to avoid laborious prompt engineering, recent works such as CoOp, CLIP-Adapter and Tip-Adapter propose to adapt vision-language models for downstream image recognition tasks on a small set of labeled data. Though promising improvements are achieved, requiring labeled data from the target datasets may restrict the scalability. In this paper, we explore a different scenario, in which the labels of the target datasets are unprovided, and we present an unsupervised prompt learning (UPL) approach to avoid prompt engineering while simultaneously improving transfer performance of CLIP-like vision-language models. As far as we know, UPL is the first work to introduce unsupervised learning into prompt learning. Experimentally, our UPL outperforms original CLIP with prompt engineering on ImageNet as well as other 10 datasets. An enhanced version of UPL is even competitive with the 8-shot CoOp and the 8-shot TIP-Adapter on most datasets. Code and models are available at \url{https://github.com/tonyhuang2022/UPL}.
\end{abstract}
\vspace{-3mm}
\section{Introduction}
\label{sec:intro}
\begin{figure}[t]
    \centering
    \begin{subfigure}[b]{0.45\textwidth}
    \centering
    \includegraphics[width=\textwidth]{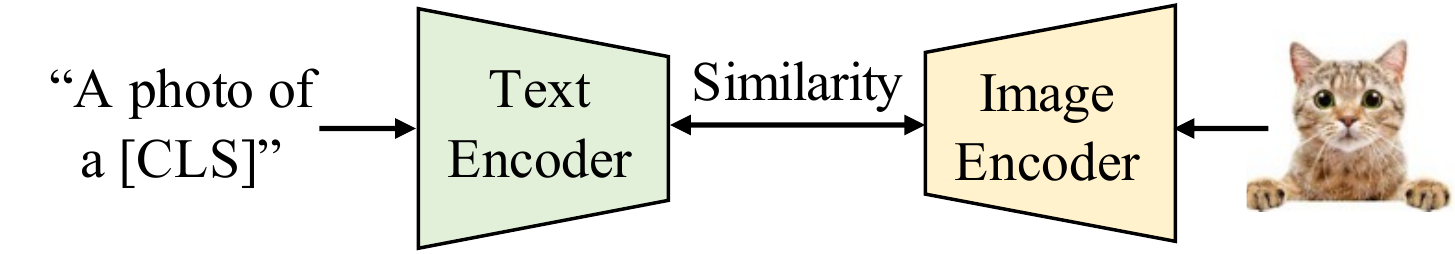}
    \vspace{-6mm}
    \caption{Inference of CLIP.}
    \label{fig:teaser_a}
    \end{subfigure}
    \hfill
    \begin{subfigure}[b]{0.23\textwidth}
    \includegraphics[width=\textwidth]{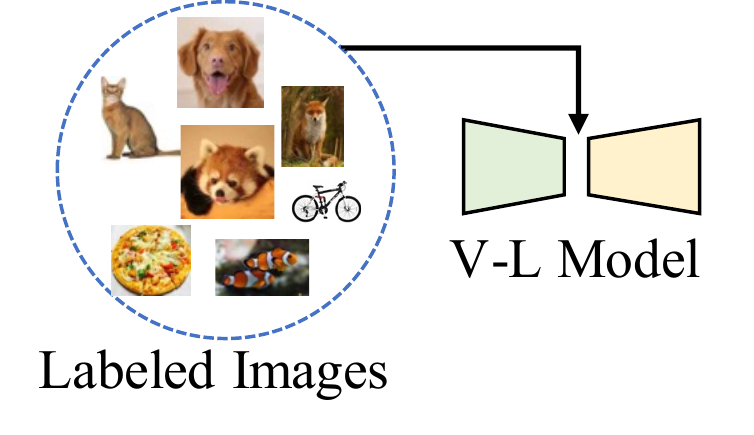}
    \vspace{-6mm}
    \caption{Existing methods adapt V-L model on labeled images.}
    \label{fig:teaser_b}
    \end{subfigure}
    \hfill
    \begin{subfigure}[b]{0.23\textwidth}
    \includegraphics[width=\textwidth]{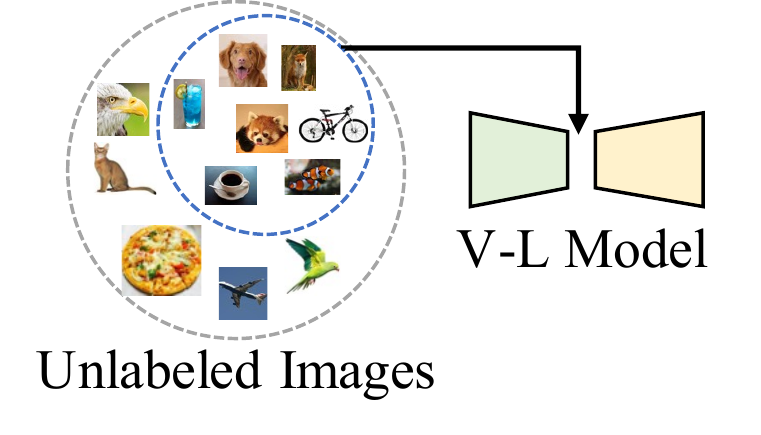}
    \vspace{-6mm}
    \caption{Ours. Adapt V-L model on selected unlabeled images.}
    \label{fig:teaser_c}
    \end{subfigure}
    \caption{(a) Inference of the pre-trained CLIP. (b) Existing methods such as CoOp, CLIP-Adapter and Tip-Adapter use a small set of \textit{labeled} images from target datasets to adapt pre-trained CLIP to downstream task. (c) Our UPL conducts prompt learning on \textit{unlabeled} images from target datasets.
    }
    \label{fig:teaser}
    \vspace{-2mm}
\end{figure}

Recently, vision-language models such as CLIP~\cite{radford2021learning}, ALIGN~\cite{jia2021scaling} and FLIP~\cite{yao2021filip} have achieved promising progress in visual representation learning and transfer learning. In contrast to traditional visual frameworks, vision-language models are trained on large-scale image-text pairs using a two-tower architecture which typically consists of an image encoder and a text encoder, to align images with raw texts in a shared embedding space. To transfer the well optimized models to downstream tasks~\cite{radford2021learning,jia2021scaling,yao2021filip,yuan2021florence,gu2021open,du2022learning,xu2021simple}, one needs to carefully design the proper text description, known as \textit{prompt}, to classify the target images correctly. For instance, one of the prompt templates used in CLIP is ``a photo of a [CLS]'' (Figure~\ref{fig:teaser_a}). However, identifying the proper prompt is non-trivial, which often requires domain knowledge and laborious prompt engineering.

To avoid hand-crafted prompt design and improve transfer performance, some supervised methods, e.g. CoOp~\cite{zhou2021learning}, CLIP-Adapter~\cite{gao2021clip} and Tip-Adapter~\cite{zhang2021tip} propose to use a small set of labeled images from the target dataset to adapt vision-language models for downstream image recognition tasks, which is illustrated in Figure~\ref{fig:teaser_b}. CoOp learns a continuous prompt representation to replace hand-crafted prompts; CLIP-Adapter adopts additional networks to learn refined features; TIP-Adapter further extends CLIP-Adapter by constructing a query-key cache model from few-shot supervisions. However, all these methods require annotated samples from the target datasets, which may limit their capacity to scale-up. In this paper, we investigate a different setting, in which the labels of target datasets are unprovided, and we propose an unsupervised prompt learning (UPL) framework to effectively adapt vision-language models for downstream image recognition task while simultaneously avoiding laborious prompt design. Figure~\ref{fig:teaser_c} shows the illustration.


Concretely, UPL firstly utilizes a pre-trained vision-language model, e.g., CLIP, to generate pseudo labels for target images, and then a self-training procedure is performed to optimize a learnable prompt representation on the selected pseudo-labeled samples. The transfer performance of the CLIP can be significantly improved by simply swapping out the hand-crafted prompt with the well-optimized prompt representation.
In contrast to threshold-based self-training methods, we select top-$K$ confident samples per class for self-training according to the observations: 1) vision-language models have biased preferences for different classes, using a pre-defined threshold to filter out unconfident samples leads to an imbalanced pseudo data distribution; 2) there is no obvious correlation between confidence scores and pseudo label accuracy, which identifies the confidence score may not be a reliable indicator to reflect the quality of pseudo labels.
Though noisy pseudo labels may be introduced simultaneously, we experimentally find that our method is robust to the noise because all classes use the same prompt representation. Inspired by the prompt ensemble strategy proposed by CLIP, we introduce pseudo label ensemble and prompt representation ensemble to further promote our methodology. Our contributions can be summarized as follows:
\vspace{-1mm}
\begin{itemize}
    \item We present an unsupervised prompt learning (UPL) framework to avoid time-consuming prompt engineering and better adapt vision-language models (e.g. CLIP) for the downstream image recognition task. As far as we know, UPL is the first work to introduce unsupervised learning into prompt learning of vision-language models.\vspace{-1mm}
\item We thoroughly analyze the characters of CLIP for pseudo-labeling. Based on the observations, we propose a series of technologies including top-$K$ pseudo-labeling strategy, pseudo label ensemble, and prompt representation ensemble to improve transfer performance.\vspace{-1mm}
\item Our UPL significantly outperforms original CLIP with prompt engineering on ImageNet and other 10 image classification datasets by large margins. An enhanced version of UPL is even competitive with supervised methods such as the 8-shot CoOp and the 8-shot TIP-Adapter on most datasets.\vspace{-1mm}
\end{itemize}
\vspace{-2mm}
\section{Related Work}
\noindent \textbf{Vision-language Models.}
Vision-language models pre-trained on large-scale image-text pairs have demonstrated great potential in visual representation learning. CLIP~\cite{radford2021learning} creates a 400 million dataset, ALIGN~\cite{jia2021scaling} exploits 1.8 billion noisy image-text pairs, FLIP~\cite{yao2021filip} collects a set of 300 million paired data for fine-grained vision-language pre-training, Wukong~\cite{gu2022wukong} presents a large-scale Chinese cross-modal dataset containing 100 million data for benchmarking different multi-modal pre-training methods, and Florence~\cite{yuan2021florence} constructs a 900 million image-text-pair dataset called FLD-900M and achieves new state-of-the-art results in majority of 44 representative benchmarks. These vision-language models all utilize a two-tower architecture consisting of a vision (image) encoder with ResNet~\cite{he2016deep}, ViT~\cite{dosovitskiy2020image} or Swin Transformer~\cite{liu2021swin}, and a language (text) encoder with standard Transformers~\cite{vaswani2017attention}. To align images with raw texts in the embedding space, text-to-image and image-to-text contrastive learning~\cite{van2018representation} are adopted.
In contrast to self-supervised pretraining approaches~\cite{grill2020bootstrap,chen2020simple,he2020momentum,chen2021exploring} for visual representation learning, vision-language models have inherent transfer capacity of image recognition. Moreover, the representative framework CLIP has been adapted to a series of vision tasks, such as object detection~\cite{gu2021open,du2022learning}, semantic segmentation~\cite{xu2021simple}, action recognition~\cite{wang2021actionclip}, video clip retrieval~\cite{luo2021clip4clip}, video caption~\cite{tang2021clip4caption} and 3D recognition~\cite{zhang2021pointclip}.

\noindent\textbf{Prompt Learning.}
Pre-trained vision-language models use prompts (e.g., ``a photo of a [CLS]'') to generate class embeddings for image recognition. Identifying the proper prompt is non-trivial, which often takes a significant amount of time for prompt engineering. Inspired by the progress of prompt learning in NLP~\cite{zhong2021factual,li2021prefix,lester2021power,shin2020autoprompt,jiang2020can}, CoOp~\cite{zhou2021learning} proposes a continuous prompt optimization strategy to avoid prompt design. CLIP-Adapter~\cite{gao2021clip} trains additional adapter networks to map the text feature and image feature to a new embedding space to better adapt the target dataset. Tip-Adapter~\cite{zhang2021tip} further extends CLIP-Adapter by creating the weights by a key-value
cache model constructed from the few-shot training set. However, all these methods rely on few-shot labeled data, which may limit their capacity
to scale-up. In contrast, our proposed UPL improves transfer performance of pre-trained vision-language models while having no requirements of annotations of the target datasets.

\noindent\textbf{Self-training.}
Self-training~\cite{scudder1965probability,yarowsky1995unsupervised,riloff1996automatically} is a simple semi-supervised learning approach. In this paradigm, a well-trained model first generates pseudo labels on the unlabeled datasets, and then the model is finetuned by using both labeled data and pseudo-labeled data. Recently, self-training has shown significant progress in deep learning, e.g., image classification~\cite{yalniz2019billion,xie2020self}, object detection~\cite{xu2021end,sohn2020simple}, semantic segmentation~\cite{hu2021semi}, speech recognition~\cite{kahn2020self,parthasarathi2019lessons}, action recognition~\cite{xu2021cross} and machine translation~\cite{he2019revisiting}. Vision-language models are usually pre-trained on large-scale image-text pairs (e.g. 400 million data for CLIP) and show promising transfer performance via prompting. Our proposed UPL generates pseudo labels for the target datasets and optimizes the continuous prompt representations via a well-designed self-training approach. This processing greatly boosts the transfer performance. In contrast to traditional self-training which finetunes all layers in a network, UPL only optimizes the continuous prompt representations while keeping the whole networks (i.e. image encoder and text encoder) fixed. As far as we know, UPL is the first work of introducing self-training into prompt learning of vision-language models.

\vspace{-2mm}
\section{Method}
\vspace{-1mm}
\begin{figure*}[t]
\centering
\includegraphics[width=0.98\linewidth]{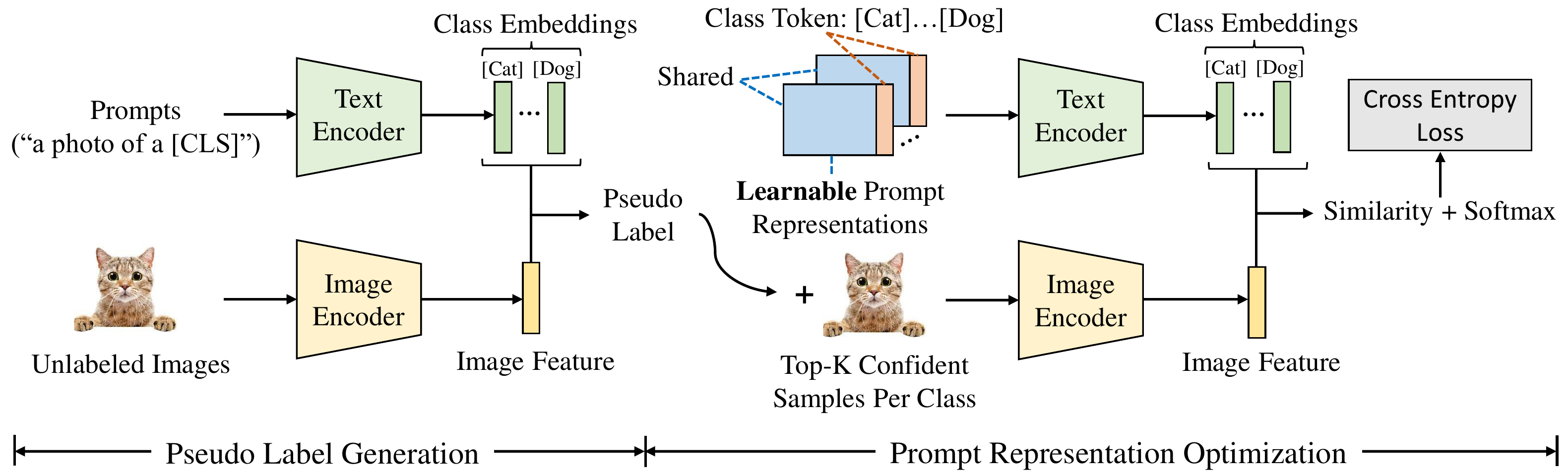}
\vspace{-2mm}
\caption{
Overview of the proposed unsupervised prompt learning (UPL) framework. Our UPL mainly contains two parts, namely pseudo label generation and prompt representation optimization. We first use CLIP with a simple prompt (e.g., ``a photo of a [CLS]'') to generate pseudo labels for target datasets and select top-$K$ confident samples per class for subsequent training. Then we define a learnable prompt representation which is optimized on selected pseudo-labeled samples. For inference, we simply swap out the hand-crafted prompts with the well-optimized prompt representations.}
\label{fig:overview}
\vspace{-2mm}
\end{figure*}

In this section, we introduce our unsupervised prompt learning (UPL) framework for vision-language models, especially for CLIP~\cite{radford2021learning}. UPL aims to avoid laborious prompt engineering while improving the transfer performance of pre-trained vision-language models. Unlike prior supervised methods~\cite{zhou2021learning,gao2021clip,zhang2021tip}, UPL does not require any annotations of the target datasets. In this section, we first present an overview of our UPL. Next, we introduce the processing of generating pseudo labels for target images. Finally, the details of prompt representation optimization via a well-designed self-training approach are described.

\vspace{-2mm}
\subsection{Overview of UPL}
\label{sec:overview}
Our UPL aims to avoid prompt engineering and boost the transfer performance of vision-language models in an unsupervised manner. Figure~\ref{fig:overview} shows an overview. UPL mainly consists of two modules, namely pseudo label generation and prompt representation optimization. In the first step, we utilize a pre-trained vision-language model (e.g. CLIP) to generate pseudo labels for unlabeled images from the target dataset.
Based on the observations that: 1) the correlation between confidence scores and pseudo label accuracy is relatively low; 2) vision-language models have biased per-class accuracy, we thus select top-$K$ confident samples for each class, instead of keeping all samples with confidence scores higher than a pre-defined threshold, for the subsequent prompt representation optimization. In the second step, we define a learnable prompt representation which is inspired by CoOp~\cite{zhou2021learning}. The prompt representation is shared across all categories and optimized on the selected unlabeled samples with generated pseudo-labels. In the transfer stage, we simply swap out the hand-crafted prompts with the well-optimized prompt representations and employ the CLIP inference pipeline for image recognition.
\vspace{-2mm}
\subsection{Pseudo Label Generation}
\label{sec:pseudo}
\noindent\textbf{Inference of CLIP.} CLIP is pre-trained on large-scale image-text pairs to align the images to the raw texts in a common embedding space. We first revisit the inference of CLIP. Given a target dataset containing $C$ classes, CLIP converts the prompt, e.g. ``a photo of a [CLS]\footnote{We use [CLS] to denote class token, and [CLS] is replaced by the specific class name, such as ``cat” or “car” in the inference.}'', into a lower-cased byte pair encoding (BPE) representation~\cite{sennrich2015neural} which is subsequently fed into the CLIP's text encoder to generate class embeddings for each category. We use $\{\boldsymbol{f}_c^{text}\}_{c=1}^C$ to denote the the set of class embeddings, where $\boldsymbol{f}_c^{text}$ denotes the class embedding of $c$-th category. Meanwhile, for an image $I$, we use $\boldsymbol{f}^{image}$ to denote its visual feature extracted by CLIP's image encoder. The probability of class $c$ is then computed as:
\begin{equation}
p_c=\frac{\exp \left(<\boldsymbol{f}_c^{text}, \boldsymbol{f}^{image}>/ \tau\right)}{\sum_{j=1}^{C} \exp \left(<\boldsymbol{f}_j^{text}, \boldsymbol{f}^{image}>/ \tau\right)},
\label{eq:prob}
\end{equation}
where $\tau$ is a temperature parameter learned by CLIP, and $<\cdot,\cdot>$ denotes cosine similarity. We can easily identify the prediction $\hat{y}$ by:
\vspace{-1mm}
\begin{equation}
\hat{y} = \argmax_c p_c,
\label{eq:pseudo_label}
\end{equation}
\vspace{-2mm}

\begin{figure}[!th]
    \centering
    \begin{subfigure}[b]{0.45\textwidth}
    \centering
    \includegraphics[width=0.7\textwidth]{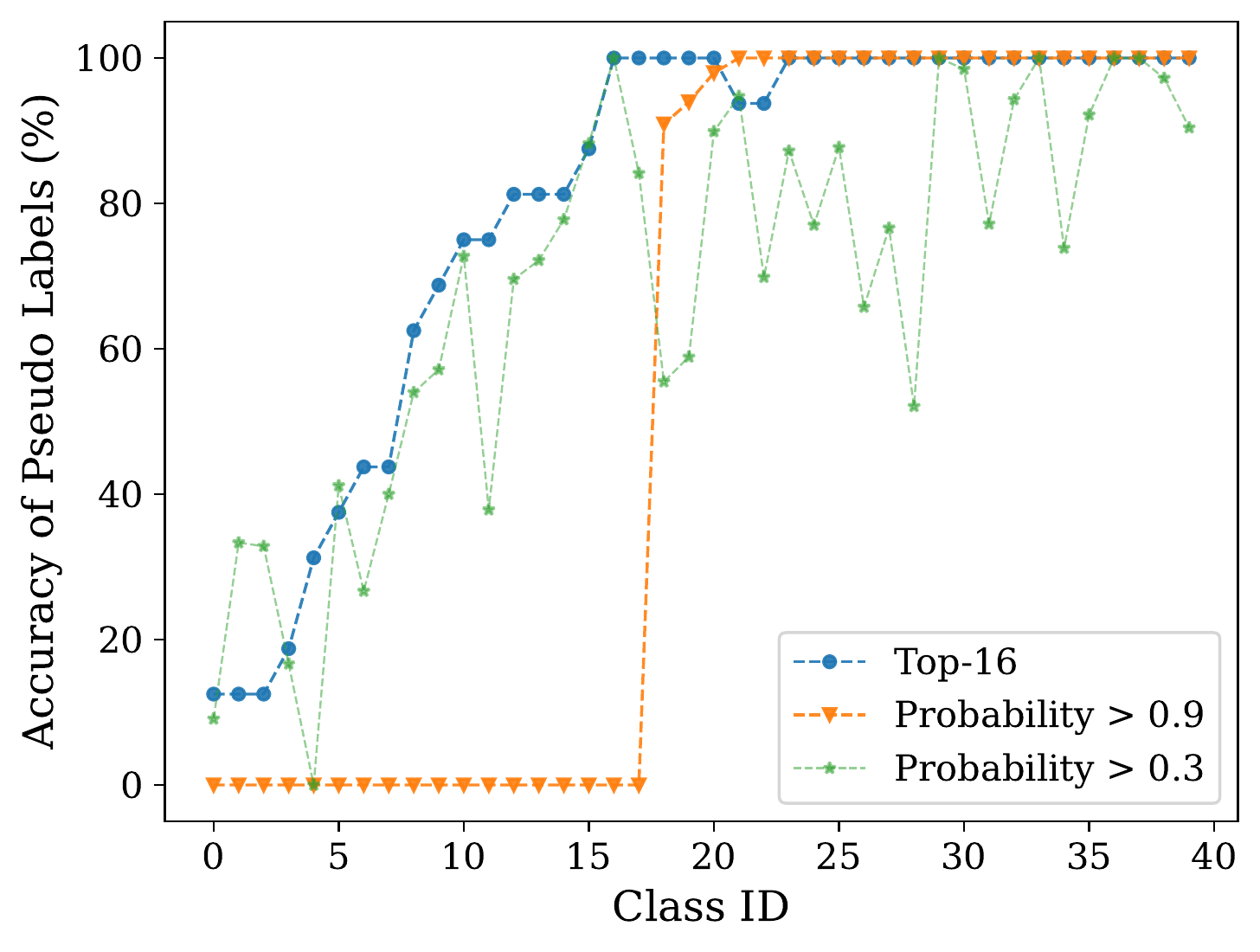}
    \vspace{-3mm}
    \caption{Per-class pseudo label accuracy of different pseudo-labeling strategies.}
    \label{fig:sampling_acc}
    \end{subfigure}
    \hfill
    \begin{subfigure}[b]{0.45\textwidth}
    \centering
    \includegraphics[width=0.7\textwidth]{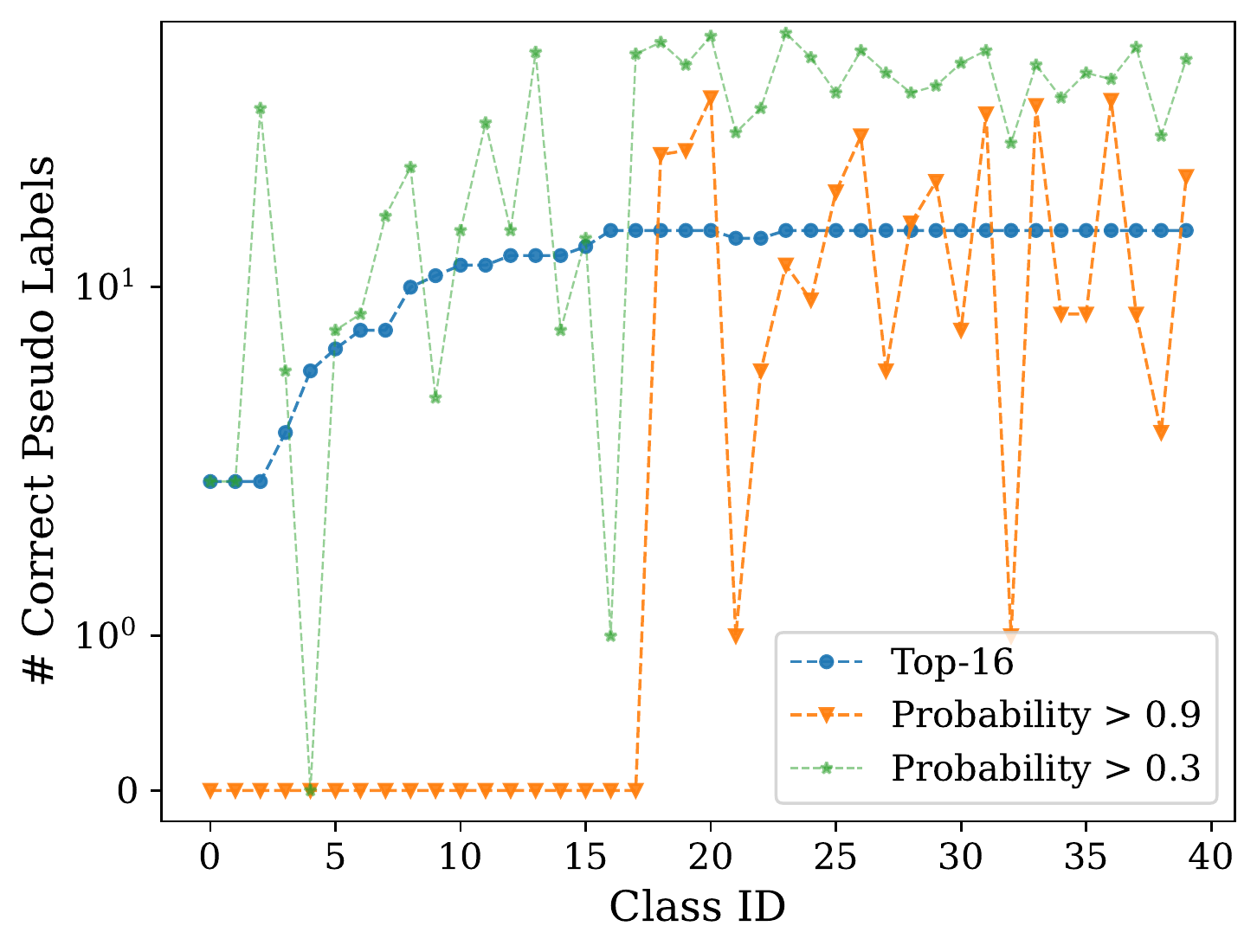}
    \vspace{-3mm}
    \caption{The number of correct pseudo labels of different pseudo-labeling strategies.}
    \label{fig:sampling_nums}
    \end{subfigure}
    \vspace{-2mm}
    \caption{Analysis of two pseudo-labeling strategies on UCF-101 dataset. We observe CLIP shows biased preferences for different classes in transfer learning. Classical self-training pre-defines a threshold to select samples of high probability, resulting in an imbalanced distribution of pseudo-labeled data (orange and green lines). We advocate to select top-$K$ confident samples per class to generate a balanced set of pseudo-labeled data for self-training (blue line). 
    }
    \label{fig:ablation_sampling}
    \vspace{-4mm}
\end{figure}

\begin{figure}[t]
\centering
\includegraphics[width=0.75\linewidth]{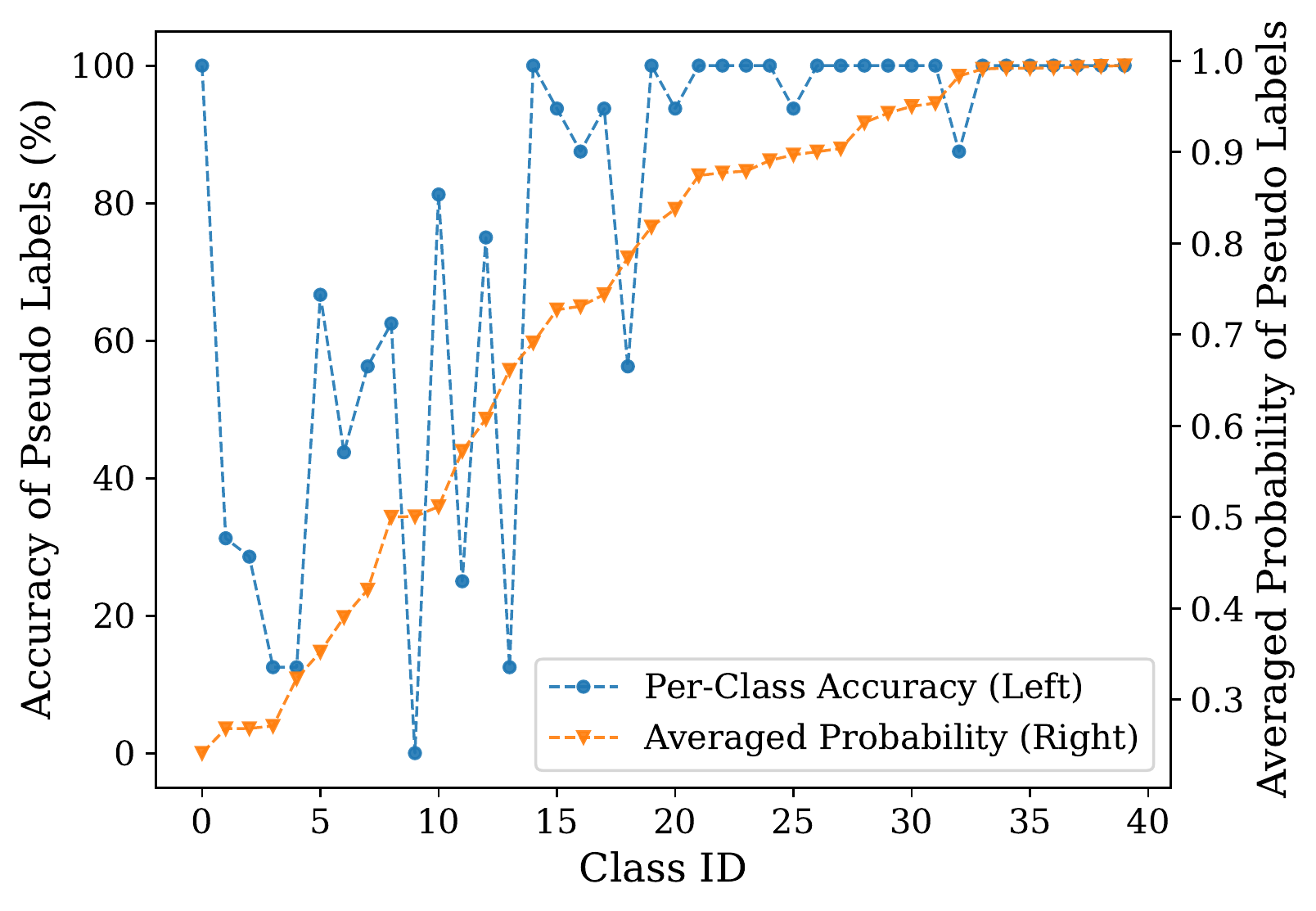}
\vspace{-4mm}
\caption{We select top-$16$ confident samples per class on UCF-101 dataset and compute averaged probability and pseudo label accuracy for each class. We observe that probability (confidence) can not completely reflect the quality of pseudo labels. It is possible for categories with low averaged probabilities to have accurate pseudo labels.
}
\vspace{-1mm}
\label{fig:pseudo_vs_acc}
\end{figure}

\noindent\textbf{Pseudo Label Generation.} Given a pre-trained CLIP, we can use Eq.\ref{eq:prob} and Eq.\ref{eq:pseudo_label} to generate pseudo labels for unlabeled samples from the target dataset. Self-training and semi-supervised learning methods usually retain the confident samples whose scores are higher than a pre-defined threshold for optimization. However, we find that it is non-trivial to directly apply this strategy on CLIP. The reasons are two-fold:
\vspace{-1mm}
\begin{itemize}
    \item We observe CLIP exhibits biased preferences for different classes when transferred to downstream image recognition task, which is mainly caused by the domain gap between the pre-training dataset and the target dataset. This phenomenon is illustrated in Figure~\ref{fig:ablation_sampling}. Using a fixed pre-defined threshold to filter out unconfident samples results in an imbalanced distribution of pseudo-labeled data, which further hinders optimization.
    \item Self-training assumes that confidence (probability) can well reflect the qualify of pseudo labels and thus a pre-defined threshold (e.g. 0.9) can be used to select high-quality samples. Nevertheless, we observe that the correlation between confidence scores and pseudo label accuracy in CLIP is comparatively weak as shown in Figure~\ref{fig:pseudo_vs_acc}.
\end{itemize}

\begin{figure}[!th]
\centering
\includegraphics[width=0.7\linewidth]{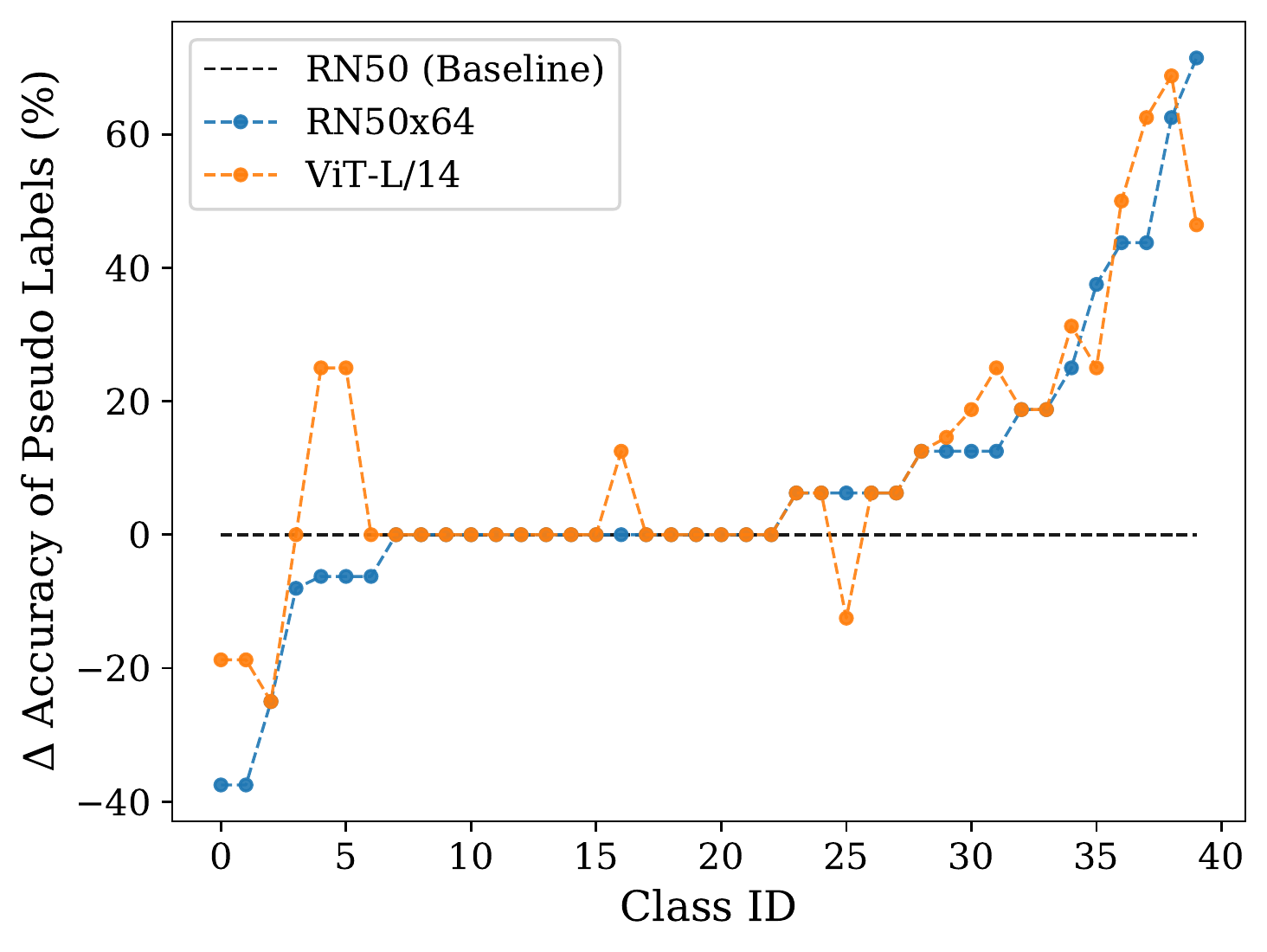}
\vspace{-3mm}
\caption{CLIP models with different vision encoders have preferences for different classes. We study this phenomenon on UCF-101 dataset. We compare three CLIP models, namely ResNet-50, ResNet-50x64 and ViT-L/14 and compute the class-wise pseudo label accuracy for each model. We show the accuracy gap between ResNet-50x64 and ResNet-50 (blue line), and the accuracy gap between ViT-L/14 and ResNet-50 (orange line).}
\vspace{-4mm}
\label{fig:model_class_aware}
\end{figure}

We therefore advocate to select top-$K$ confident samples per class by Eq.~\ref{eq:prob} and Eq.~\ref{eq:pseudo_label} for subsequent optimization. This prevents the vast number of samples of certain categories from overwhelming the model during training. We set $K=16$ experimentally.

\noindent\textbf{Pseudo Label Ensemble.} CLIP provides a series of vision models including ResNet-50, ResNet-101, ResNet-50x4, ResNet-50x16, ResNet-50x64, ViT-B/32, ViT-B/16 and ViT-L/14. We observe that CLIP models with different vision architectures have biased class-wise accuracy as shown in Figure~\ref{fig:model_class_aware}. Motivated by this finding, we propose a simple pseudo label ensemble strategy to further improve the quality of pseudo labels. In particular, given $M$ CLIP models with various vision architectures, we utilize Eq.\ref{eq:prob} to obtain probability $p_i^m$ predicted by the $m$-th CLIP model, then the final probability $\overline{p}_i$ is simply averaged by $\overline{p}_i= \sum_{m=1}^{M} p_i^m / M$. Similarly, Eq.\ref{eq:pseudo_label} can be applied on $\overline{p}_i$ to generate enhanced pseudo labels. Once the process is finished, we use pseudo-labeled data for unsupervised prompt representation optimization.

\subsection{Prompt Representation Optimization}
\label{sec:prompt_opt}
The original CLIP defines various prompt templates, e.g. ``a photo of a [CLS]'' for transfer learning. However, identifying the proper prompt is non-trivial, since it often requires domain knowledge and laborious prompt engineering. A slight change in prompt may lead to a huge difference in performance. CoOp~\cite{zhou2021learning} proposes to avoid hand-crafted prompt design by optimizing a continuous prompt representation on a small set of labeled data. Our UPL resembles CoOp, however, our method does not require any annotations of the target datasets.

\begin{table*}[!th]
\centering
\begin{tabular}{@{}l|c|cc|ccc|ccc@{}}
\toprule
        Method                  & \multicolumn{1}{c|}{CLIP}     & \multicolumn{1}{c}{UPL}    & UPL* 
                          & \multicolumn{3}{c|}{CoOp}  & \multicolumn{3}{c}{Tip-Adapter} \\   \cmidrule{1-10}
                           Prompt learning & \multicolumn{1}{c|}{-}         & \multicolumn{1}{c}{Unsupervised} & Unsupervised & \textbf{2}-shot &  \textbf{4}-shot & \textbf{8}-shot &\textbf{2}-shot & \textbf{4}-shot & \textbf{8}-shot \\ \midrule

ImageNet                  & 60.34   &  60.51  &  61.09   & 57.13 & 59.72 & 61.52 & 60.96 &60.98 &  61.45                                           \\
Caltech101                & 86.09   &  89.94    &  91.40 & 87.76 & 89.67 & 90.14 & 89.25 &89.41 &  89.94                                                  \\
DTD                       & 41.61   &   46.57   &  55.08 & 47.48 & 54.19 & 58.65 & 49.76 &54.14 &  57.33                                                 \\
EuroSAT                   & 38.23   &  54.83    &  71.04 & 59.98 & 62.17 & 68.73 & 61.10  &65.30 &   66.89                                               \\
FGVCAircraft              & 16.92   &  17.34    &  21.75 & 20.36 & 22.10 & 24.99 & 21.25 &21.54 &  24.48                                                 \\
Food101                   & 77.33   &   77.58   &  77.93 & 72.92 & 73.74 & 76.28 & 77.58 &77.60 &  77.79                                                 \\
Flowers102                & 66.06   &   68.90   &  76.65 & 76.58 & 84.59 & 88.27 & 76.82 &81.53 &  85.95                                                 \\
OxfordPets                & 85.83   &   88.28   &  89.51 & 84.53 & 87.11 & 87.71 & 87.38 &87.67 &  87.87                                                 \\
SUN397                    & 60.18   &   63.98   &  66.42 & 61.35 & 65.08 & 67.47 & 62.82 &64.32 &  65.57                                                \\
StandfordCars              & 55.64  &   62.13   &  70.97 & 59.49 & 61.92 & 65.25 & 59.86 &62.03 &  63.35                                                \\
UCF101                    & 62.70   &   67.17   &  70.18 & 65.06 & 68.26 & 71.67 & 66.59 &67.51 &  69.10                                                 \\
\midrule
\textbf{Average}          & {\textbf{59.18}} & {\textbf{63.38}} &{\textbf{68.37}}& \textbf{62.97}  & \textbf{66.23}   &   \textbf{69.15} &  \textbf{64.85} & \textbf{66.55} & \textbf{68.16}           \\ \bottomrule
\end{tabular}
\vspace{-2mm}
\caption{Main Results of UPL and UPL* on 11 datasets. We compare our unsupervised approach with: 1) original CLIP with prompt engineering~\cite{radford2021learning}; 2) supervised methods including CoOp~\cite{zhou2021learning} and Tip-Adapter~\cite{zhang2021tip}. Both UPL and UPL* boost the performance of the original CLIP with prompt engineering. UPL outperforms 2-shot CoOp. UPL* is competitive with 8-shot CoOp and 8-shot Tip-Adapter on most datasets.}
\label{Table:SOTA}
\vspace{-2mm}
\end{table*}

\noindent\textbf{Learnable Prompt Representation.} Our goal is to learn a prompt representation on pseudo-labeled data to improve CLIP's transfer performance. Formally, we define the learnable prompt representation $\boldsymbol{V} \in \mathcal{R}^{D \times L}$, where $D$ denotes the dimension of word embeddings (512 for CLIP), $L$ is a hyper-parameter which is set to 16 by default. Given a target dataset containing $C$ classes, we define the continuous prompt $\boldsymbol{V}_c \in \mathcal{R}^{D \times (L+1)}$  for class $c~(1 \leq c \leq C)$ as:
\vspace{-1mm}
\begin{equation}
\boldsymbol{V}_c=[\boldsymbol{V},\boldsymbol{w}_c],
\label{eq:prompt_def}
\end{equation}
where $\boldsymbol w_{c} \in \mathcal{R}^D$ represents the fixed word embedding of class $c$. Note that all classes share the identical prompt representation $\boldsymbol{V}$. The training is extremely simple as shown in Figure~\ref{fig:overview} (right part). For each pseudo labeled image, we extract its visual feature $\boldsymbol{f}^{image}$ by feeding the input image into CLIP's vision encoder; meanwhile, class embeddings can be generated by feeding $\{\boldsymbol{V}_c\}_{c=1}^{C}$ into the CLIP's text encoder $g(\cdot)$. The probability of $c$-th class is computed as
\begin{equation}
p_c=\frac{\exp \left(<g(\boldsymbol{V}_c), \boldsymbol{f}^{image}>/ \tau\right)}{\sum_{j=1}^{C} \exp \left(<g(\boldsymbol{V}_j), \boldsymbol{f}^{image}>/ \tau\right)},
\label{eq:UPL_prob}
\end{equation}
where $\tau$ is the temperature parameter. For a training image, we calculate the probabilities of all classes by Eq.~\ref{eq:UPL_prob} and minimize the cross-entropy loss with its pseudo label. The gradients can back-propagate through the text encoder $g(\cdot)$, which takes advantage of the wealth of information encoded in the text encoder, and finally update the learnable prompt representation $\boldsymbol{V}$. Notice the weights of both image encoder and text encoder remain unchanged during training.

\noindent\textbf{Inference.} Once the optimization of prompt representation $\boldsymbol{V}$ is finished, given the target dataset, we feed $\{\boldsymbol{V}_c\}_{c=1}^{C}$ into the CLIP's text encoder to generate class embeddings for all categories. For a test image, we simply feed it into the CLIP's image encoder to extract its visual feature and apply Eq.\ref{eq:UPL_prob} to compute the probabilities for image recognition.

\noindent\textbf{Prompt Representation Ensemble.} Original CLIP defines numerous prompts to enhance its transfer performance, inspiring us to learn multi prompt representations with diverse initializations. Concretely, we independently optimize $N$ randomly initialized prompt representations. In the inference stage, we compute the probabilities predicted by all prompt representations and the averaged probability is served as the final prediction.

\label{sec:inference}

\section{Experiment}

\subsection{Implementation Details}
\label{experiment_settings}

\noindent\textbf{Vision-language Models.} 
We use CLIP~\cite{radford2021learning} as our pre-trained vision-language model. UPL applied on CLIP with ResNet-50~\cite{he2016deep} is served as our baseline. As illustrated in Figure~\ref{fig:model_class_aware}, we observe that CLIP models with different vision encoders have preferences for different categories. Therefore, we propose an enhanced version named UPL* which leverages additional CLIP models with various vision architectures including ResNet-101, ResNet50x4, ResNet50x16, ResNet50x64, ViT-B/32, ViT-B/16 and ViT-L/14~\cite{dosovitskiy2020image} to improve the quality of pseudo labels. Note that these additional models are merely used for pseudo-labeling, UPL* still uses the same network architecture as UPL (CLIP with ResNet-50).

\noindent\textbf{Pseudo Label Generation.} 
CLIP designs a series of prompt templates for inference, e.g., 80 hand-crafted prompts for ImageNet. Involving all prompts provided by CLIP for pseudo label generation may go against our desire of avoiding laborious prompt engineering. Thus, we only use the simplest prompt to generate pseudo labels. For example, we adopt promot ``a photo of a [CLS]'' on ImageNet. Please refer to supplementary material for more details about prompt templates used in pseudo label generation. Unless otherwise specific, we select top-$16$ confident samples per class to optimize prompt representation.

\noindent\textbf{Learnable Prompt Representations.}
The prompt representations are randomly initialized by drawing from a zero-mean Gaussian distribution with standard deviation equal to 0.02. We set length $L=16$ in Eq.\ref{eq:prompt_def} by default. We use $16$ prompt representations for ensemble in system-level comparison with prior methods. Unless otherwise specific, for all ablation studies, we use a single prompt representation for efficiency.

\noindent\textbf{Training Details.}
We use SGD with an initial learning rate of 0.002 and a cosine decay learning rate
scheduler for optimization. We train 50 epochs for all datasets, and the batch size is set as 32. We warm up the training in the first epoch with a fixed learning rate of 1e-5.

\noindent\textbf{Datasets.}
Following CLIP~\cite{radford2021learning} and CoOp~\cite{zhou2021learning}, we use 11 publicly available image classification datasets including ImageNet~\cite{deng2009imagenet}, Caltech101~\cite{fei2004learning}, DTD~\cite{cimpoi2014describing}, EuroSAT~\cite{helber2019eurosat}, FGVCAircraft~\cite{maji2013fine}, Food101~\cite{bossard2014food}, Flowers102~\cite{nilsback2008automated}, OxfordPets~\cite{parkhi2012cats}, SUN397~\cite{xiao2010sun}, StandfordCars~\cite{krause20133d}, and UCF101~\cite{soomro2012ucf101}. These datasets cover a variety of different visual classification tasks, such as general objects, fine-grained and even textures classification, constituting a comprehensive benchmark.

\vspace{-1mm}
\subsection{Main Results}
\label{performance_analysis}
The main results across 11 datasets are reported in Table \ref{Table:SOTA}. We compare our approach with: 1) the original CLIP with prompt engineering; 2) supervised methods including CoOp~\cite{zhou2021learning} and Tip-Adapter~\cite{zhang2021tip}. Original CLIP defines a number of prompt templates (e.g. 80 for ImageNet) to improve the transfer performance. Our UPL not only avoids such prompt engineering, but also outperforms CLIP by $+4.2$ point of the averaged accuracy. Our UPL*, which involves different CLIP models for pseudo-labeling while using the single CLIP with ResNet-50 for inference, further boosts the averaged accuracy to $68.37$. Compared with supervised methods, UPL surpasses 2-shot CoOp, and UPL* is competitive with 8-shot CoOp and 8-shot Tip-Adapter on most datasets (ImageNet, Caltech101, EuroSAT, Food101, OxfordPets, SUN397 and StandfordCars).

\begin{table}[!t]
    \centering
    \begin{tabular}{l|c}
    \toprule
        Strategy & Accuracy  \\ 
        \midrule
        Confidence$>$ 0.3 & 63.83 \\ 
        Confidence$>$ 0.9 & 57.32 \\ 
        Top-16 (default) & \textbf{64.84} \\ \bottomrule
    \end{tabular}
    \vspace{-1mm}
        \caption{Ablation study of different pseudo-labeling strategies on UCF101 dataset. We compare our top-$K$ strategy with confidence threshold strategies which are commonly used in self-training and semi-supervised learning.}
    \label{ablation:sample_strategy}
    \vspace{-1mm}
\end{table}

\begin{figure}[!t] 
\centering
\includegraphics[width=0.7\linewidth]{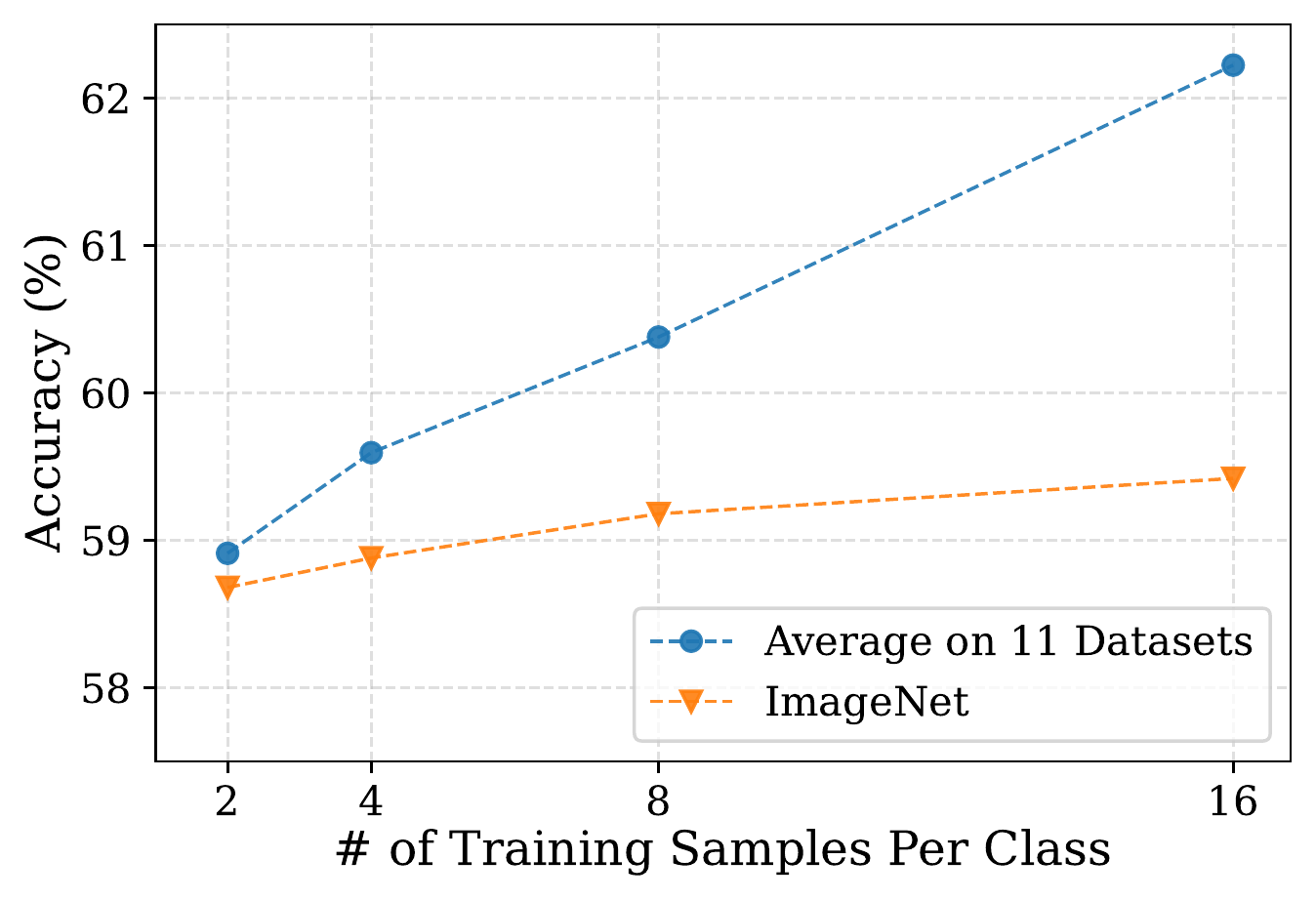}
\vspace{-1mm}
\caption{Ablation study on top-$K$ pseudo-labeling strategy with varying $K$. With the increase of pseudo-labeled samples, the performance increases.}
\label{fig:nums_shots}
\vspace{-1mm}
\end{figure}

\subsection{Ablation Study}
\label{saming_strategies} 
\noindent\textbf{Different Pseudo-labeling Strategies.} 
Traditional self-training and semi-supervised learning approaches usually select unlabeled samples whose confidence scores are higher than a pre-defined threshold as pseudo-labeled data. Here we compare our proposed top-$K$ pseudo-labeling strategy with threshold-based strategies in Table~\ref{ablation:sample_strategy}. We also visualize the pseudo label accuracy and the number of correct pseudo labels for different strategies in Figure~\ref{fig:ablation_sampling}. Using a higher threshold (0.9) causes an imbalanced distribution of pseudo-labeled data, while a lower threshold (0.3) introduces too much noise which disturbs the training. In contrast, our default top-$K$ pseudo-labeling strategy ensures a balanced distribution of pseudo-labeled data and prevent the vast number of samples of certain categories from overwhelming the model during training.

\noindent\textbf{Select Top-$K$ Confident Samples as Pseudo-labeled Data.} 
We have shown the superiority of top-$K$ pseudo-labeling strategy, here we study how many pseudo-labeled samples ($K$) of each class should be used. Concretely, we vary the value of $K$ ($K=2,4,8,16$) and train our UPL with different amount of pseudo-labeled data. The results are visualized in Figure~\ref{fig:nums_shots}. With the increase of pseudo-labeled samples, the performance increases. Thus we set $K=16$.

\begin{table}[t]
    \centering

    \begin{tabular}{c|c|c|c}
    \toprule
        \multirow{2}{*}{Dataset} & \multirow{2}{*}{Model} & \multicolumn{2}{c}{Accuracy}  \\ \cmidrule{3-4} 
         &  & Pseudo label & Transfer \\ \midrule
       \multirow{4}{*}{UCF101}      & ResNet-50           & 79.34 & 64.84 \\
                                    & ResNet-50x64        & 88.38 & 68.71 \\
                                    & ViT-B/14      & 90.50 & 67.72 \\
                                    & Ensemble   & \textbf{90.51} & \textbf{68.86}\\ \midrule

        \multirow{4}{*}{SUN397}     & ResNet-50     & 78.87 & 62.83 \\
                                    & ResNet-50x64  & 83.47 & 64.87 \\
                                    & ViT-B/14      & 83.95 & 64.77 \\
                                    & Ensemble    & \textbf{87.18} & \textbf{65.02}\\ \bottomrule
    \end{tabular}
    \vspace{-2mm}
         \caption{Ablation study of pseudo label ensemble strategy on UCF101 and SUN397 datasets. For UPL*, we use different CLIP models with various vision encoders to generate pseudo labels and evaluate the pseudo label accuracy as well as the transfer accuracy. Notice that different CLIP models are only used for pseudo-labeling, the training and inference are still performed on CLIP with ResNet-50.}
          \vspace{-1mm}
    \label{ablation:model_ensemble}
\end{table}

\begin{table}[!t]
    \centering
    \begin{subtable}{0.3\linewidth}
        \centering
        \begin{tabular}{c|c}
            \toprule
             $L$ & Acc  \\ \midrule
            4 & 89.81 \\ 
            8 & 89.58 \\
            16 & 89.79 \\ \bottomrule
        \end{tabular}
        \vspace{-1mm}
        \caption{Caltech101.}
    \end{subtable}
    \begin{subtable}{0.3\linewidth}
        \centering
        \begin{tabular}{c|c}
            \toprule
             $L$ & Acc \\ \midrule
            4 & 46.02 \\ 
            8 & 46.28 \\
            16 & 45.98 \\ \bottomrule
        \end{tabular}
        \vspace{-1mm}
        \caption{DTD.}
    \end{subtable}
    \begin{subtable}{0.3\linewidth}
        \centering
        \begin{tabular}{c|c}
            \toprule
             $L$ & Acc  \\ \midrule
            4 & 59.36 \\ 
            8 & 60.61 \\
            16 & 60.75 \\ \bottomrule
        \end{tabular}
        \vspace{-1mm}
        \caption{StandfordCars.}
    \end{subtable}
    \vspace{-3mm}
    \caption{Study the effects of length $L$ of prompt representation on Caltech101, DTD and StandfordCars. UPL is less sensitive to the change of $L$.}
    \label{ablation:prompt_representation_length}
    \vspace{-6mm}
\end{table}

\noindent\textbf{Pseudo Label Ensemble.}
As illustrated in Figure~\ref{fig:model_class_aware}, CLIP models with various vision architectures have preferences for different classes. Here we quantitatively evaluate the pseudo label accuracy as well as the the transfer accuracy of using different CLIP models for pseudo-labeling. Table~\ref{ablation:model_ensemble} shows the results.

\noindent\textbf{The Length of Prompt Representation.}
We study the length $L$ of the prompt representation in Table~\ref{ablation:prompt_representation_length}.

\begin{figure}[t]
    \centering
    \begin{subfigure}[b]{0.45\textwidth}
         \centering
         \includegraphics[width=0.7\textwidth]{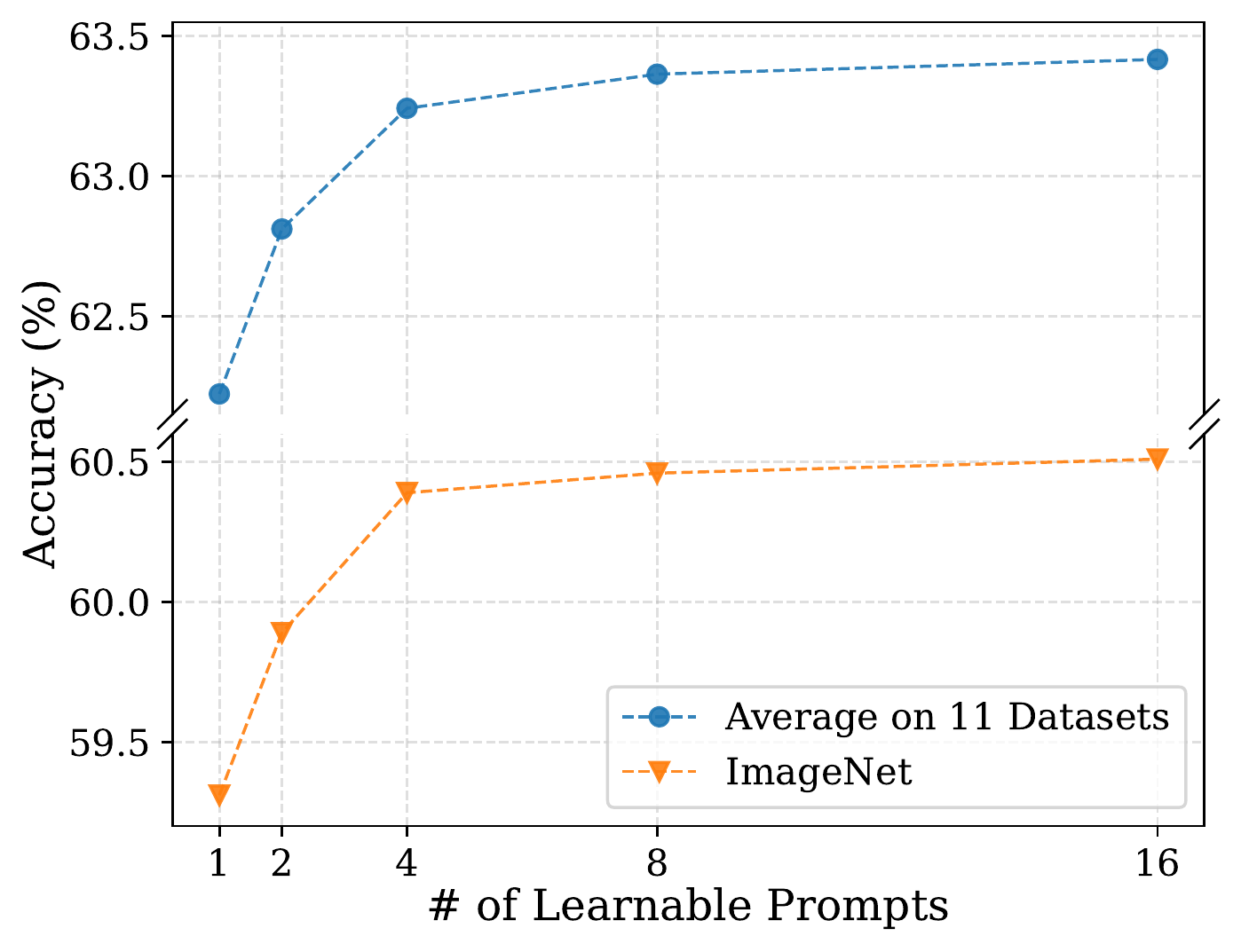}
          \vspace{-2mm}
         \caption{Study of ensembling different numbers of prompt representations.}
         \label{fig:nums_prompts}
     \end{subfigure}
     \hfill
     \begin{subfigure}[b]{0.45\textwidth}
         \centering
         \includegraphics[width=0.7\textwidth]{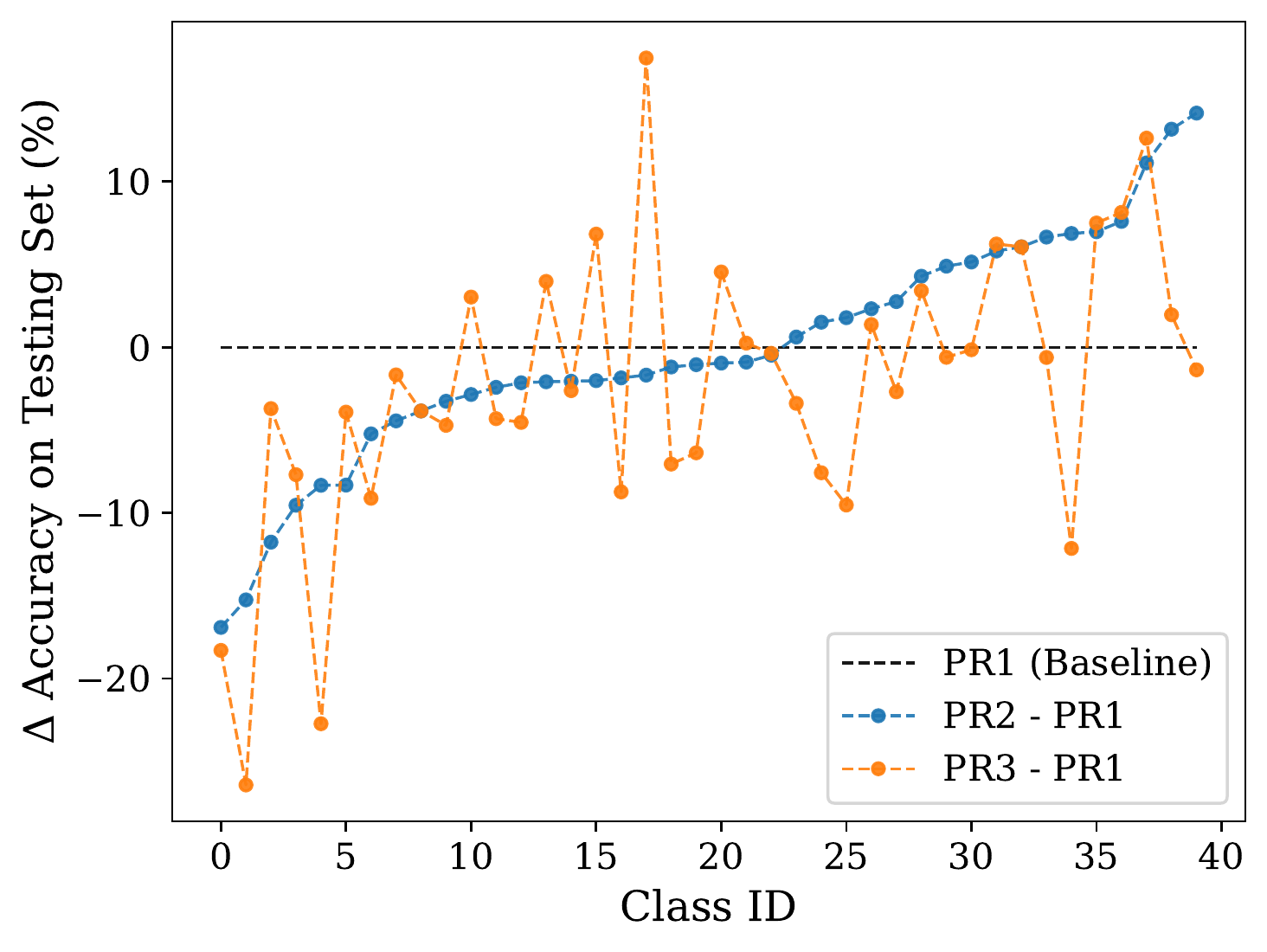}
         \vspace{-2mm}
         \caption{Well-learned prompt representations show preferences for different classes.}
         \label{fig:prompt_class_aware}
     \end{subfigure}
     \vspace{-3mm}
    \caption{(a) We study the effects of ensembling different numbers of the prompt representations on UCF101 dataset. (b) We optimize three prompt representations (PR-1, 2 and 3) with different initializations on UCF101 dataset. We use PR-1 as baseline and calculate the per-class accuracy difference with PR-2 and PR-3. We find the well-learned prompt representations have biased preferences for different classes, inspiring us to ensemble them to facilitate transfer performance.}
    \label{fig:ablation_nums}
    \vspace{-5mm}
\end{figure}

\noindent\textbf{Prompt Representation Ensemble.} 
CLIP designs a series of prompt templates to promote transfer performance, inspiring us to learn multi prompt representations and ensemble them in the inference stage. Here we study the effects of ensembling different numbers ($N=2,4,8,16$) of the prompt representations in Figure~\ref{fig:nums_prompts}. We find the performance is almost saturated when $N=16$. Next, we investigate the effectiveness of the prompt representation ensemble. To be specific, we independently optimize three prompt representations (PR-1, 2 and 3) using our UPL on the UCF101 dataset. For each of the three well learned prompt representations, we compute the per-class accuracy on UCF101 test set. We use PR-1 as our baseline and calculate per-class accuracy difference with the rest two (PR-2 and PR-3). The results are visualized in Figure~\ref{fig:prompt_class_aware}. Though the overall accuracy of three prompt representations are almost 
same (64.58, 64.66, 64.39 for PR-1, 2 and 3 respectively), the per-class accuracy differs significantly. Thus prompt representation ensemble promotes transfer.

\begin{figure}[t]
\centering
\includegraphics[width=0.7\linewidth]{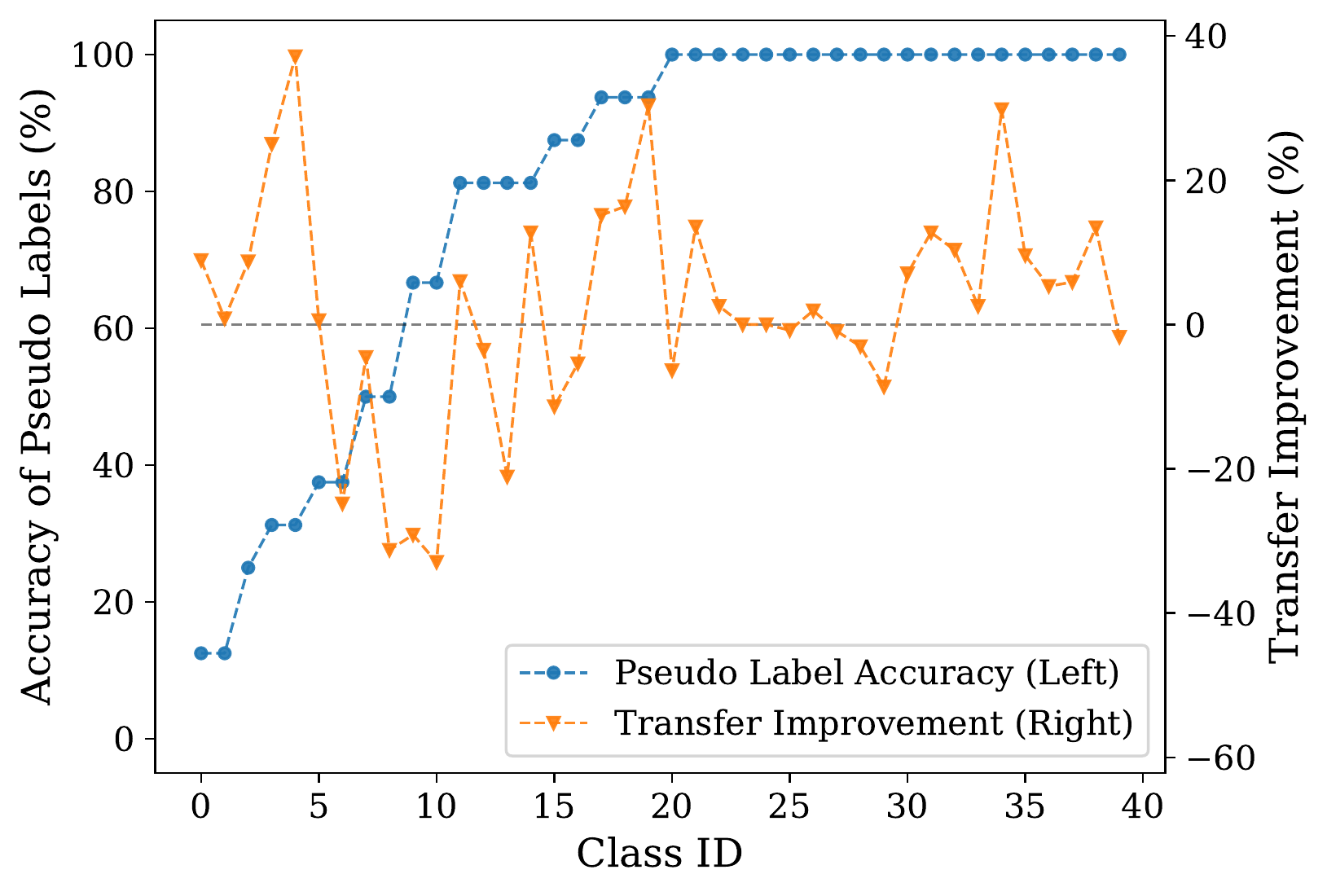}
 \vspace{-4mm}
\caption{We calculate per-class pseudo label accuracy and per-class transfer improvement on UCF101 dataset. There are no obvious correspondences between pseudo label accuracy and transfer accuracy. We still see significant transfer improvement for these classes with low pseudo label accuracy, indicating our UPL is robust to the noisy pseudo-labeled samples.}
 \vspace{-4mm}
\label{fig:acc_vs_conf}
\end{figure}

\noindent\textbf{Robustness to Noisy Pseudo Labels.} Pseudo labels play an important role in self-training. The crucial principle of pseudo-labeling is to generate a sufficient amount of pseudo-labeled samples with high quality. Some recent works~\cite{hu2021semi,rizve2021defense,xu2021end} have explored the directions of reducing negative impacts of noisy pseudo labels. As shown in Figure~\ref{fig:sampling_acc}, CLIP has biased per-class accuracy, indicating that the pseudo labels predicted by CLIP are noisy for the under-performing categories. It is natural to raise a question: are there any correspondences between per-class pseudo label accuracy and per-class transfer accuracy? To answer this question, we perform the following calculations on the UCF101 dataset: 1) per-class pseudo label accuracy; 2) per-class transfer improvement (accuracy difference between UPL and original CLIP). Figure~\ref{fig:acc_vs_conf} shows the results. We observe that there are no obvious correspondences existing between pseudo label accuracy and transfer accuracy. In fact, we still observe significant transfer improvement for some categories with low pseudo label accuracy, showing that our UPL is robust to pseudo label noise. Since all classes share the identical prompt representation (see Eq.~\ref{eq:prompt_def}), the prompt representations are optimized on the pseudo-labeled samples from all classes. Though noisy pseudo labels exist in some certain categories, we can nevertheless optimize a favourable shared prompt representation on a large amount of qualified pseudo-labeled samples.
\vspace{-3mm}
\section{Conclusion}
\vspace{-1mm}
In this paper, we propose an unsupervised prompt learning (UPL) framework to avoid time-consuming prompt engineering and simultaneously facilitate the transfer performance of CLIP. In contrast to prior supervised approaches such as CoOp, CLIP-Adapter, and TIP-Adapter, our UPL is the first unsupervised framework to better adapt pre-trained vision-language models to the downstream image recognition task. We conduct abundant experiments on ImageNet as well as 10 widely used image classification datasets. Our UPL outperforms the original CLIP with prompt engineering on Imagenet and other 10 datasets. Furthermore, our unsupervised method also outperforms 2-shot CoOp in terms of the averaged accuracy across 11 datasets, and an improved version of UPL is even competitive with the 8-shot CoOp and the 8-shot Tip-Adapter on most datasets.
\clearpage
\bibliography{egbib}

\begin{thebibliography}{55}
\providecommand{\natexlab}[1]{#1}

\bibitem[{nil(2008)}]{nilsback2008automated}
 2008.
\newblock Automated flower classification over a large number of classes.
\newblock In \emph{2008 Sixth Indian Conference on Computer Vision, Graphics \&
  Image Processing}, 722--729. IEEE.

\bibitem[{Bossard, Guillaumin, and Van~Gool(2014)}]{bossard2014food}
Bossard, L.; Guillaumin, M.; and Van~Gool, L. 2014.
\newblock Food-101--mining discriminative components with random forests.
\newblock In \emph{European conference on computer vision}, 446--461. Springer.

\bibitem[{Chen et~al.(2020)Chen, Kornblith, Norouzi, and
  Hinton}]{chen2020simple}
Chen, T.; Kornblith, S.; Norouzi, M.; and Hinton, G. 2020.
\newblock A simple framework for contrastive learning of visual
  representations.
\newblock In \emph{International conference on machine learning}, 1597--1607.
  PMLR.

\bibitem[{Chen and He(2021)}]{chen2021exploring}
Chen, X.; and He, K. 2021.
\newblock Exploring simple siamese representation learning.
\newblock In \emph{Proceedings of the IEEE/CVF Conference on Computer Vision
  and Pattern Recognition}, 15750--15758.

\bibitem[{Cimpoi et~al.(2014)Cimpoi, Maji, Kokkinos, Mohamed, and
  Vedaldi}]{cimpoi2014describing}
Cimpoi, M.; Maji, S.; Kokkinos, I.; Mohamed, S.; and Vedaldi, A. 2014.
\newblock Describing textures in the wild.
\newblock In \emph{Proceedings of the IEEE Conference on Computer Vision and
  Pattern Recognition}, 3606--3613.

\bibitem[{Deng et~al.(2009)Deng, Dong, Socher, Li, Li, and
  Fei-Fei}]{deng2009imagenet}
Deng, J.; Dong, W.; Socher, R.; Li, L.-J.; Li, K.; and Fei-Fei, L. 2009.
\newblock Imagenet: A large-scale hierarchical image database.
\newblock In \emph{2009 IEEE conference on computer vision and pattern
  recognition}, 248--255. Ieee.

\bibitem[{Dosovitskiy et~al.(2020)Dosovitskiy, Beyer, Kolesnikov, Weissenborn,
  Zhai, Unterthiner, Dehghani, Minderer, Heigold, Gelly
  et~al.}]{dosovitskiy2020image}
Dosovitskiy, A.; Beyer, L.; Kolesnikov, A.; Weissenborn, D.; Zhai, X.;
  Unterthiner, T.; Dehghani, M.; Minderer, M.; Heigold, G.; Gelly, S.; et~al.
  2020.
\newblock An image is worth 16x16 words: Transformers for image recognition at
  scale.
\newblock \emph{arXiv preprint arXiv:2010.11929}.

\bibitem[{Du et~al.(2022)Du, Wei, Zhang, Shi, Gao, and Li}]{du2022learning}
Du, Y.; Wei, F.; Zhang, Z.; Shi, M.; Gao, Y.; and Li, G. 2022.
\newblock Learning to Prompt for Open-Vocabulary Object Detection with
  Vision-Language Model.
\newblock \emph{arXiv preprint arXiv:2203.14940}.

\bibitem[{Fei-Fei, Fergus, and Perona(2004)}]{fei2004learning}
Fei-Fei, L.; Fergus, R.; and Perona, P. 2004.
\newblock Learning generative visual models from few training examples: An
  incremental bayesian approach tested on 101 object categories.
\newblock In \emph{2004 conference on computer vision and pattern recognition
  workshop}, 178--178. IEEE.

\bibitem[{Gao et~al.(2021)Gao, Geng, Zhang, Ma, Fang, Zhang, Li, and
  Qiao}]{gao2021clip}
Gao, P.; Geng, S.; Zhang, R.; Ma, T.; Fang, R.; Zhang, Y.; Li, H.; and Qiao, Y.
  2021.
\newblock Clip-adapter: Better vision-language models with feature adapters.
\newblock \emph{arXiv preprint arXiv:2110.04544}.

\bibitem[{Grill et~al.(2020)Grill, Strub, Altch{\'e}, Tallec, Richemond,
  Buchatskaya, Doersch, Avila~Pires, Guo, Gheshlaghi~Azar
  et~al.}]{grill2020bootstrap}
Grill, J.-B.; Strub, F.; Altch{\'e}, F.; Tallec, C.; Richemond, P.;
  Buchatskaya, E.; Doersch, C.; Avila~Pires, B.; Guo, Z.; Gheshlaghi~Azar, M.;
  et~al. 2020.
\newblock Bootstrap your own latent-a new approach to self-supervised learning.
\newblock \emph{Advances in Neural Information Processing Systems}, 33:
  21271--21284.

\bibitem[{Gu et~al.(2022)Gu, Meng, Lu, Hou, Niu, Xu, Liang, Zhang, Jiang, and
  Xu}]{gu2022wukong}
Gu, J.; Meng, X.; Lu, G.; Hou, L.; Niu, M.; Xu, H.; Liang, X.; Zhang, W.;
  Jiang, X.; and Xu, C. 2022.
\newblock Wukong: 100 Million Large-scale Chinese Cross-modal Pre-training
  Dataset and A Foundation Framework.
\newblock \emph{arXiv preprint arXiv:2202.06767}.

\bibitem[{Gu et~al.(2021)Gu, Lin, Kuo, and Cui}]{gu2021open}
Gu, X.; Lin, T.-Y.; Kuo, W.; and Cui, Y. 2021.
\newblock Open-vocabulary Object Detection via Vision and Language Knowledge
  Distillation.
\newblock \emph{arXiv preprint arXiv:2104.13921}.

\bibitem[{He et~al.(2019)He, Gu, Shen, and Ranzato}]{he2019revisiting}
He, J.; Gu, J.; Shen, J.; and Ranzato, M. 2019.
\newblock Revisiting self-training for neural sequence generation.
\newblock \emph{arXiv preprint arXiv:1909.13788}.

\bibitem[{He et~al.(2020)He, Fan, Wu, Xie, and Girshick}]{he2020momentum}
He, K.; Fan, H.; Wu, Y.; Xie, S.; and Girshick, R. 2020.
\newblock Momentum contrast for unsupervised visual representation learning.
\newblock In \emph{Proceedings of the IEEE/CVF conference on computer vision
  and pattern recognition}, 9729--9738.

\bibitem[{He et~al.(2016)He, Zhang, Ren, and Sun}]{he2016deep}
He, K.; Zhang, X.; Ren, S.; and Sun, J. 2016.
\newblock Deep residual learning for image recognition.
\newblock In \emph{Proceedings of the IEEE conference on computer vision and
  pattern recognition}, 770--778.

\bibitem[{Helber et~al.(2019)Helber, Bischke, Dengel, and
  Borth}]{helber2019eurosat}
Helber, P.; Bischke, B.; Dengel, A.; and Borth, D. 2019.
\newblock Eurosat: A novel dataset and deep learning benchmark for land use and
  land cover classification.
\newblock \emph{IEEE Journal of Selected Topics in Applied Earth Observations
  and Remote Sensing}, 12(7): 2217--2226.

\bibitem[{Hu et~al.(2021)Hu, Wei, Hu, Ye, Cui, and Wang}]{hu2021semi}
Hu, H.; Wei, F.; Hu, H.; Ye, Q.; Cui, J.; and Wang, L. 2021.
\newblock Semi-Supervised Semantic Segmentation via Adaptive Equalization
  Learning.
\newblock \emph{Advances in Neural Information Processing Systems}, 34.

\bibitem[{Jia et~al.(2021)Jia, Yang, Xia, Chen, Parekh, Pham, Le, Sung, Li, and
  Duerig}]{jia2021scaling}
Jia, C.; Yang, Y.; Xia, Y.; Chen, Y.-T.; Parekh, Z.; Pham, H.; Le, Q.; Sung,
  Y.-H.; Li, Z.; and Duerig, T. 2021.
\newblock Scaling up visual and vision-language representation learning with
  noisy text supervision.
\newblock In \emph{International Conference on Machine Learning}, 4904--4916.
  PMLR.

\bibitem[{Jiang et~al.(2020)Jiang, Xu, Araki, and Neubig}]{jiang2020can}
Jiang, Z.; Xu, F.~F.; Araki, J.; and Neubig, G. 2020.
\newblock How can we know what language models know?
\newblock \emph{Transactions of the Association for Computational Linguistics},
  8: 423--438.

\bibitem[{Kahn, Lee, and Hannun(2020)}]{kahn2020self}
Kahn, J.; Lee, A.; and Hannun, A. 2020.
\newblock Self-training for end-to-end speech recognition.
\newblock In \emph{ICASSP 2020-2020 IEEE International Conference on Acoustics,
  Speech and Signal Processing (ICASSP)}, 7084--7088. IEEE.

\bibitem[{Krause et~al.(2013)Krause, Stark, Deng, and Fei-Fei}]{krause20133d}
Krause, J.; Stark, M.; Deng, J.; and Fei-Fei, L. 2013.
\newblock 3d object representations for fine-grained categorization.
\newblock In \emph{Proceedings of the IEEE international conference on computer
  vision workshops}, 554--561.

\bibitem[{Lester, Al-Rfou, and Constant(2021)}]{lester2021power}
Lester, B.; Al-Rfou, R.; and Constant, N. 2021.
\newblock The power of scale for parameter-efficient prompt tuning.
\newblock \emph{arXiv preprint arXiv:2104.08691}.

\bibitem[{Li and Liang(2021)}]{li2021prefix}
Li, X.~L.; and Liang, P. 2021.
\newblock Prefix-tuning: Optimizing continuous prompts for generation.
\newblock \emph{arXiv preprint arXiv:2101.00190}.

\bibitem[{Liu et~al.(2021)Liu, Lin, Cao, Hu, Wei, Zhang, Lin, and
  Guo}]{liu2021swin}
Liu, Z.; Lin, Y.; Cao, Y.; Hu, H.; Wei, Y.; Zhang, Z.; Lin, S.; and Guo, B.
  2021.
\newblock Swin transformer: Hierarchical vision transformer using shifted
  windows.
\newblock In \emph{Proceedings of the IEEE/CVF International Conference on
  Computer Vision}, 10012--10022.

\bibitem[{Luo et~al.(2021)Luo, Ji, Zhong, Chen, Lei, Duan, and
  Li}]{luo2021clip4clip}
Luo, H.; Ji, L.; Zhong, M.; Chen, Y.; Lei, W.; Duan, N.; and Li, T. 2021.
\newblock Clip4clip: An empirical study of clip for end to end video clip
  retrieval.
\newblock \emph{arXiv preprint arXiv:2104.08860}.

\bibitem[{Maji et~al.(2013)Maji, Rahtu, Kannala, Blaschko, and
  Vedaldi}]{maji2013fine}
Maji, S.; Rahtu, E.; Kannala, J.; Blaschko, M.; and Vedaldi, A. 2013.
\newblock Fine-grained visual classification of aircraft.
\newblock \emph{arXiv preprint arXiv:1306.5151}.

\bibitem[{Parkhi et~al.(2012)Parkhi, Vedaldi, Zisserman, and
  Jawahar}]{parkhi2012cats}
Parkhi, O.~M.; Vedaldi, A.; Zisserman, A.; and Jawahar, C. 2012.
\newblock Cats and dogs.
\newblock In \emph{2012 IEEE conference on computer vision and pattern
  recognition}, 3498--3505. IEEE.

\bibitem[{Parthasarathi and Strom(2019)}]{parthasarathi2019lessons}
Parthasarathi, S. H.~K.; and Strom, N. 2019.
\newblock Lessons from building acoustic models with a million hours of speech.
\newblock In \emph{ICASSP 2019-2019 IEEE International Conference on Acoustics,
  Speech and Signal Processing (ICASSP)}, 6670--6674. IEEE.

\bibitem[{Radford et~al.(2021)Radford, Kim, Hallacy, Ramesh, Goh, Agarwal,
  Sastry, Askell, Mishkin, Clark et~al.}]{radford2021learning}
Radford, A.; Kim, J.~W.; Hallacy, C.; Ramesh, A.; Goh, G.; Agarwal, S.; Sastry,
  G.; Askell, A.; Mishkin, P.; Clark, J.; et~al. 2021.
\newblock Learning transferable visual models from natural language
  supervision.
\newblock In \emph{International Conference on Machine Learning}, 8748--8763.
  PMLR.

\bibitem[{Riloff(1996)}]{riloff1996automatically}
Riloff, E. 1996.
\newblock Automatically generating extraction patterns from untagged text.
\newblock In \emph{Proceedings of the national conference on artificial
  intelligence}, 1044--1049.

\bibitem[{Rizve et~al.(2021)Rizve, Duarte, Rawat, and Shah}]{rizve2021defense}
Rizve, M.~N.; Duarte, K.; Rawat, Y.~S.; and Shah, M. 2021.
\newblock In defense of pseudo-labeling: An uncertainty-aware pseudo-label
  selection framework for semi-supervised learning.
\newblock \emph{arXiv preprint arXiv:2101.06329}.

\bibitem[{Scudder(1965)}]{scudder1965probability}
Scudder, H. 1965.
\newblock Probability of error of some adaptive pattern-recognition machines.
\newblock \emph{IEEE Transactions on Information Theory}, 11(3): 363--371.

\bibitem[{Sennrich, Haddow, and Birch(2015)}]{sennrich2015neural}
Sennrich, R.; Haddow, B.; and Birch, A. 2015.
\newblock Neural machine translation of rare words with subword units.
\newblock \emph{arXiv preprint arXiv:1508.07909}.

\bibitem[{Shin et~al.(2020)Shin, Razeghi, Logan~IV, Wallace, and
  Singh}]{shin2020autoprompt}
Shin, T.; Razeghi, Y.; Logan~IV, R.~L.; Wallace, E.; and Singh, S. 2020.
\newblock Autoprompt: Eliciting knowledge from language models with
  automatically generated prompts.
\newblock \emph{arXiv preprint arXiv:2010.15980}.

\bibitem[{Sohn et~al.(2020)Sohn, Zhang, Li, Zhang, Lee, and
  Pfister}]{sohn2020simple}
Sohn, K.; Zhang, Z.; Li, C.-L.; Zhang, H.; Lee, C.-Y.; and Pfister, T. 2020.
\newblock A simple semi-supervised learning framework for object detection.
\newblock \emph{arXiv preprint arXiv:2005.04757}.

\bibitem[{Soomro, Zamir, and Shah(2012)}]{soomro2012ucf101}
Soomro, K.; Zamir, A.~R.; and Shah, M. 2012.
\newblock UCF101: A dataset of 101 human actions classes from videos in the
  wild.
\newblock \emph{arXiv preprint arXiv:1212.0402}.

\bibitem[{Tang et~al.(2021)Tang, Wang, Liu, Rao, Li, and
  Li}]{tang2021clip4caption}
Tang, M.; Wang, Z.; Liu, Z.; Rao, F.; Li, D.; and Li, X. 2021.
\newblock CLIP4Caption: CLIP for Video Caption.
\newblock In \emph{Proceedings of the 29th ACM International Conference on
  Multimedia}, 4858--4862.

\bibitem[{Van~den Oord, Li, and Vinyals(2018)}]{van2018representation}
Van~den Oord, A.; Li, Y.; and Vinyals, O. 2018.
\newblock Representation learning with contrastive predictive coding.
\newblock \emph{arXiv e-prints}, arXiv--1807.

\bibitem[{Vaswani et~al.(2017)Vaswani, Shazeer, Parmar, Uszkoreit, Jones,
  Gomez, Kaiser, and Polosukhin}]{vaswani2017attention}
Vaswani, A.; Shazeer, N.; Parmar, N.; Uszkoreit, J.; Jones, L.; Gomez, A.~N.;
  Kaiser, {\L}.; and Polosukhin, I. 2017.
\newblock Attention is all you need.
\newblock \emph{Advances in neural information processing systems}, 30.

\bibitem[{Wang, Xing, and Liu(2021)}]{wang2021actionclip}
Wang, M.; Xing, J.; and Liu, Y. 2021.
\newblock Actionclip: A new paradigm for video action recognition.
\newblock \emph{arXiv preprint arXiv:2109.08472}.

\bibitem[{Xiao et~al.(2010)Xiao, Hays, Ehinger, Oliva, and
  Torralba}]{xiao2010sun}
Xiao, J.; Hays, J.; Ehinger, K.~A.; Oliva, A.; and Torralba, A. 2010.
\newblock Sun database: Large-scale scene recognition from abbey to zoo.
\newblock In \emph{2010 IEEE computer society conference on computer vision and
  pattern recognition}, 3485--3492. IEEE.

\bibitem[{Xie et~al.(2020)Xie, Luong, Hovy, and Le}]{xie2020self}
Xie, Q.; Luong, M.-T.; Hovy, E.; and Le, Q.~V. 2020.
\newblock Self-training with noisy student improves imagenet classification.
\newblock In \emph{Proceedings of the IEEE/CVF conference on computer vision
  and pattern recognition}, 10687--10698.

\bibitem[{Xu et~al.(2021{\natexlab{a}})Xu, Zhang, Hu, Wang, Wang, Wei, Bai, and
  Liu}]{xu2021end}
Xu, M.; Zhang, Z.; Hu, H.; Wang, J.; Wang, L.; Wei, F.; Bai, X.; and Liu, Z.
  2021{\natexlab{a}}.
\newblock End-to-end semi-supervised object detection with soft teacher.
\newblock In \emph{Proceedings of the IEEE/CVF International Conference on
  Computer Vision}, 3060--3069.

\bibitem[{Xu et~al.(2021{\natexlab{b}})Xu, Zhang, Wei, Lin, Cao, Hu, and
  Bai}]{xu2021simple}
Xu, M.; Zhang, Z.; Wei, F.; Lin, Y.; Cao, Y.; Hu, H.; and Bai, X.
  2021{\natexlab{b}}.
\newblock A Simple Baseline for Zero-shot Semantic Segmentation with
  Pre-trained Vision-language Model.
\newblock \emph{arXiv preprint arXiv:2112.14757}.

\bibitem[{Xu et~al.(2021{\natexlab{c}})Xu, Wei, Sun, Yang, Shen, Dai, Zhou, and
  Lin}]{xu2021cross}
Xu, Y.; Wei, F.; Sun, X.; Yang, C.; Shen, Y.; Dai, B.; Zhou, B.; and Lin, S.
  2021{\natexlab{c}}.
\newblock Cross-Model Pseudo-Labeling for Semi-Supervised Action Recognition.
\newblock \emph{arXiv preprint arXiv:2112.09690}.

\bibitem[{Yalniz et~al.(2019)Yalniz, J{'e}gou, Chen, Paluri, and
  Mahajan}]{yalniz2019billion}
Yalniz, I.~Z.; J{'e}gou, H.; Chen, K.; Paluri, M.; and Mahajan, D. 2019.
\newblock Billion-scale semi-supervised learning for image classification.
\newblock \emph{arXiv preprint arXiv:1905.00546}.

\bibitem[{Yao et~al.(2021)Yao, Huang, Hou, Lu, Niu, Xu, Liang, Li, Jiang, and
  Xu}]{yao2021filip}
Yao, L.; Huang, R.; Hou, L.; Lu, G.; Niu, M.; Xu, H.; Liang, X.; Li, Z.; Jiang,
  X.; and Xu, C. 2021.
\newblock FILIP: Fine-grained Interactive Language-Image Pre-Training.
\newblock \emph{arXiv preprint arXiv:2111.07783}.

\bibitem[{Yarowsky(1995)}]{yarowsky1995unsupervised}
Yarowsky, D. 1995.
\newblock Unsupervised word sense disambiguation rivaling supervised methods.
\newblock In \emph{33rd annual meeting of the association for computational
  linguistics}, 189--196.

\bibitem[{Yuan et~al.(2021)Yuan, Chen, Chen, Codella, Dai, Gao, Hu, Huang, Li,
  Li et~al.}]{yuan2021florence}
Yuan, L.; Chen, D.; Chen, Y.-L.; Codella, N.; Dai, X.; Gao, J.; Hu, H.; Huang,
  X.; Li, B.; Li, C.; et~al. 2021.
\newblock Florence: A New Foundation Model for Computer Vision.
\newblock \emph{arXiv preprint arXiv:2111.11432}.

\bibitem[{Zhang et~al.(2021{\natexlab{a}})Zhang, Fang, Gao, Zhang, Li, Dai,
  Qiao, and Li}]{zhang2021tip}
Zhang, R.; Fang, R.; Gao, P.; Zhang, W.; Li, K.; Dai, J.; Qiao, Y.; and Li, H.
  2021{\natexlab{a}}.
\newblock Tip-Adapter: Training-free CLIP-Adapter for Better Vision-Language
  Modeling.
\newblock \emph{arXiv preprint arXiv:2111.03930}.

\bibitem[{Zhang et~al.(2021{\natexlab{b}})Zhang, Guo, Zhang, Li, Miao, Cui,
  Qiao, Gao, and Li}]{zhang2021pointclip}
Zhang, R.; Guo, Z.; Zhang, W.; Li, K.; Miao, X.; Cui, B.; Qiao, Y.; Gao, P.;
  and Li, H. 2021{\natexlab{b}}.
\newblock PointCLIP: Point Cloud Understanding by CLIP.
\newblock \emph{arXiv preprint arXiv:2112.02413}.

\bibitem[{Zhong, Friedman, and Chen(2021)}]{zhong2021factual}
Zhong, Z.; Friedman, D.; and Chen, D. 2021.
\newblock Factual probing is [mask]: Learning vs. learning to recall.
\newblock \emph{arXiv preprint arXiv:2104.05240}.

\bibitem[{Zhou et~al.(2021)Zhou, Yang, Loy, and Liu}]{zhou2021learning}
Zhou, K.; Yang, J.; Loy, C.~C.; and Liu, Z. 2021.
\newblock Learning to prompt for vision-language models.
\newblock \emph{arXiv preprint arXiv:2109.01134}.

\bibitem[{Zhou et~al.(2022)Zhou, Yang, Loy, and Liu}]{zhou2022conditional}
Zhou, K.; Yang, J.; Loy, C.~C.; and Liu, Z. 2022.
\newblock Conditional prompt learning for vision-language models.
\newblock In \emph{Proceedings of the IEEE/CVF Conference on Computer Vision
  and Pattern Recognition}, 16816--16825.

\end{thebibliography}
\clearpage
\appendix
\section{More Experiments}
\subsection{Generalization of UPL}
We explore our UPL with the CoOp-style~\cite{zhou2021learning} structure in the main paper due to its simplicity, however, the proposed unsupervised paradigm (UPL) can be adapted to any prompt learning methods as an alternative of the few-shot paradigm. Table~\ref{R2_tuning_strategies} shows the results of applying UPL to different frameworks including CoCoOp~\cite{zhou2022conditional}, CLIP-Adapter~\cite{gao2021clip} and Tip-Adapter-F~\cite{zhang2021tip}. All methods equipped with UPL outperform the original CLIP~\cite{radford2021learning}, showing the generalization of our method.

\begin{table}[!h]
    \centering
    \begin{tabular}{l|c|c|c}
    \toprule
        Method & Caltech101 & UCF101 & DTD  \\ \midrule
        Original CLIP & 86.09 & 62.70 & 41.61 \\
        \midrule
        UPL w/ CoOp & 89.94  & 67.17 & 46.57  \\
        UPL w/ CoCoOp & 90.47 & 66.80 & 46.63  \\
        UPL w/ CLIP-Adapter & 87.71  & 63.18 & 44.80 \\ 
        UPL w/ Tip-Adapter-F & 89.78 & 64.26 & 46.75  \\ 
    \bottomrule 
    \end{tabular}
    \caption{Apply UPL to different frameworks.}
    \label{R2_tuning_strategies}
\end{table}

\subsection{Position of the [CLS] Token}
In the main paper, the [CLS] token is positioned in the end of the prompt representation as defined in Eq.3 of the main paper. Here we explore the different positions of the [CLS] token and report the results in Table~\ref{R3_prompt_position}. Three different ways achieve similar performance, which demonstrates the proposed UPL is robust to the position of [CLS] token.

\begin{table}[!h]
    \centering
    
    \begin{tabular}{l|c|c|c}
    \toprule
        Position & Caltech101 & UCF 101 & DTD   \\ \midrule
        Original CLIP & 86.09 & 62.70 & 41.61 \\
        \midrule
        Frontal  & 89.98 & 66.48   & 46.99  \\
        Middle   & 90.43 &  66.75  & 46.51  \\
        End      & 89.94 &  67.17  & 46.57  \\ \bottomrule
    \end{tabular}
    \caption{Study of inserting the [CLS] token into different positions.}
    \label{R3_prompt_position}
\end{table}

\subsection{Comparison with CLIP of Single Prompt}
The results of CLIP~\cite{radford2021learning} on ImageNet and other 10 datasets are obtained by fusing the predictions of multiple prompts in Table~1 of the main paper. Here we compare our method with CLIP in single-prompt scenario. The results are shown in Table~\ref{R2_clip_one_prompt}. Our UPL demonstrates its superiority over the original CLIP in both one-prompt and multiple-prompt scenarios.

\begin{table}[!h]
    \centering
    \setlength{\tabcolsep}{4pt}
    
    \begin{tabular}{l|c|c|c|c}
    \toprule
       Method & \#Prompt & ImageNet  & Caltech101  & UCF101 \\ 
       \midrule
       CLIP & 1 & 58.21  & 85.92  & 58.29 \\
       UPL & 1 & 59.30  & 89.12  & 64.70 \\
       \midrule
       CLIP & Multiple & 60.34 (80)  & 86.09 (34)  & 62.70 (48) \\
       UPL & Multiple & 60.51 (16)  & 89.94 (16)  & 67.17 (16) \\
    \bottomrule 
    \end{tabular}
    \caption{Study on UPL/CLIP with different number of prompts. We show the prompt number used in UPL and CLIP in parentheses.}
    \label{R2_clip_one_prompt}
\end{table}

\section{Visualization Results}
\subsection{Nearest Words of Prompt Representations}

\begin{table*}[!t]
    \centering
    \begin{tabular}{c|lc|lc|lc|lc|lc}
    \toprule
      ID & \multicolumn{2}{c|}{ImageNet}  & \multicolumn{2}{c|}{Food101} &  \multicolumn{2}{c|}{OxfordPets} & \multicolumn{2}{c|}{DTD}  & \multicolumn{2}{c}{UCF101} \\ \midrule 
    1& grp&1.20 & kc&0.53 & milo&1.85 & kle&0.80 & chillin&0.86 \\
    2& shows&0.94 & led&0.52 & pi&1.09 & cutout&0.79 & N/A & 0.74\\
    3& beh&1.07 & ila&0.53 & calls&1.30 & con&0.72 & u&0.86 \\
    4& b&0.92 & detri&0.56 & N/A & 1.12 & boston&0.79 & alter&0.63 \\
    5& listing&1.27 & 2&0.49 & cat&1.32 & favorite&0.63 & criti&0.66 \\
    6& on&1.03 & pie&0.59 & *&0.83 & ist&0.86 & starting&0.80 \\
    7& did&0.91 & join&0.63 & zes&1.29 & bod&0.67 & presents&0.82 \\
    8& N/A & 0.96 & a&0.62 & radi&2.38 & roe&1.06 & ranked&0.98 \\
    9& then&1.04 & N/A & 0.57 & sey&2.30 & spring&0.90 & pattern&1.04 \\
    10& it&0.98 & over&0.63 & usage&1.29 & acmilan&0.93 & hookah&0.77 \\
    11& tweeter&1.12 & ...:&0.64 & courtesy&1.17 & dusk&0.79 & decorate&0.85 \\
    12& ds&1.19 & at&0.60 & xx&1.28 & nap&0.69 & siddi&1.00 \\
    13& exp&1.10 & on&0.56 & nad&1.18 & celebrations&0.86 & ates&0.83 \\
    14& held&1.19 & N/A & 0.64 & gp&1.48 & they&0.87 & coming&0.74 \\
    15& N/A & 0.99 & yemen&0.74 & N/A & 2.10 & N/A & 1.00 & ayyy&1.09 \\
    16& a&1.26 & minute&0.58 & 3&1.24 & ley&1.05 & only&1.07 \\ \bottomrule
    \end{tabular}
    \caption{We search within the vocabulary for words that are closest to the well-optimized representations learned by UPL according to the Euclidean distance on ImageNet, Food101, OxfordPets, DTD, and UCF101. N/A denotes non-Latin characters.}
    \label{words_different_datasets}
\end{table*}

\begin{table*}[!h]
    \centering
    \begin{tabular}{c|lc|lc|lc|lc}
    \toprule
      ID & \multicolumn{2}{c|}{PR-1}  & \multicolumn{2}{c|}{PR-2} &  \multicolumn{2}{c|}{PR-3} & \multicolumn{2}{c}{PR-4}  \\ 
      \midrule 
        1& grp&1.20 & rt&1.24 & algorithms&1.15 & relaxing&1.43 \\
        2& shows&0.94 & around&0.89 & zana&1.36 & azu&1.15 \\
        3& beh&1.07 & once&1.11 & dome&1.09 & .(&1.27 \\
        4& b&0.92 & N/A & 0.88 & much&1.13 & N/A & 1.18 \\
        5& listing&1.27 & than&0.76 & !!!!!!!!!!&0.99 & hanging&1.11 \\
        6& on&1.03 & trajec&1.08 & now&0.78 & look&1.01 \\
        7& did&0.91 & N/A & 0.91 & i&1.01 & nin&1.16 \\
        8& N/A & 0.96 & \& &1.20 & N/A & 0.99 & 2&1.10 \\
        9& then&1.04 & dt&0.98 & cant&0.86 & unit&0.94 \\
        10& it&0.98 & N/A & 0.99 & thepersonalnetwork&1.07 & probab&1.04 \\
        11& tweeter&1.12 & aw&0.91 & ,&1.17 & ard&0.78 \\
        12& ds&1.19 & pushing&0.96 & pride&1.18 & :...&1.10 \\
        13& exp&1.10 & whom&1.02 & -(&0.92 & epp&1.08 \\
        14& held&1.19 & t&0.98 & who&1.26 & no&1.11 \\
        15& N/A & 0.99 & a&1.46 & todo&1.12 & a&1.19 \\
        16& a&1.26 & a&1.16 & a&1.11 & ;;&1.06 \\
     \bottomrule
    \end{tabular}
        \caption{We show the nearest words of four well-optimized prompt representations (PR-1, 2, 3 and 4) on ImageNet. N/A denotes non-Latin characters.}
    \label{words_different_prompts}
\end{table*}

Interpreting the well-optimized prompt representations is not easy since they are optimized in a continuous space. Following CoOp~\cite{zhou2021learning}, we search within the vocabulary for words that are closest to the optimized prompt representations (each prompt representation consists of 16 learnable vectors) based on the Euclidean distance. Note that CLIP~\cite{radford2021learning} utilizes the BPE representation~\cite{sennrich2015neural} for tokenization, so the vocabulary includes subwords that frequently appear in the text. We first compare the nearest words of well-optimized prompt representations trained by our UPL for five datasets (ImageNet, Food101, OxfordPets, DTD and UCF101) in Table~\ref{words_different_datasets}. Though we observe no obvious correspondences existing among optimized prompt representations and subwords, it is surprising to find that some nearest subwords show very strong correlations with the corresponding dataset, such as ``pie'' for Food101, ``cat'' for OxfordPets. Then, we report the nearest words of four well-optimized prompts with different initializations on ImageNet in Table~\ref{words_different_prompts} to explore the relationship among different optimized prompt representations. 

\subsection{Visualization of Predictions}

\begin{figure*}[h]
     \centering
     \begin{subfigure}[b]{0.48\linewidth}
         \centering
         \includegraphics[width=\textwidth]{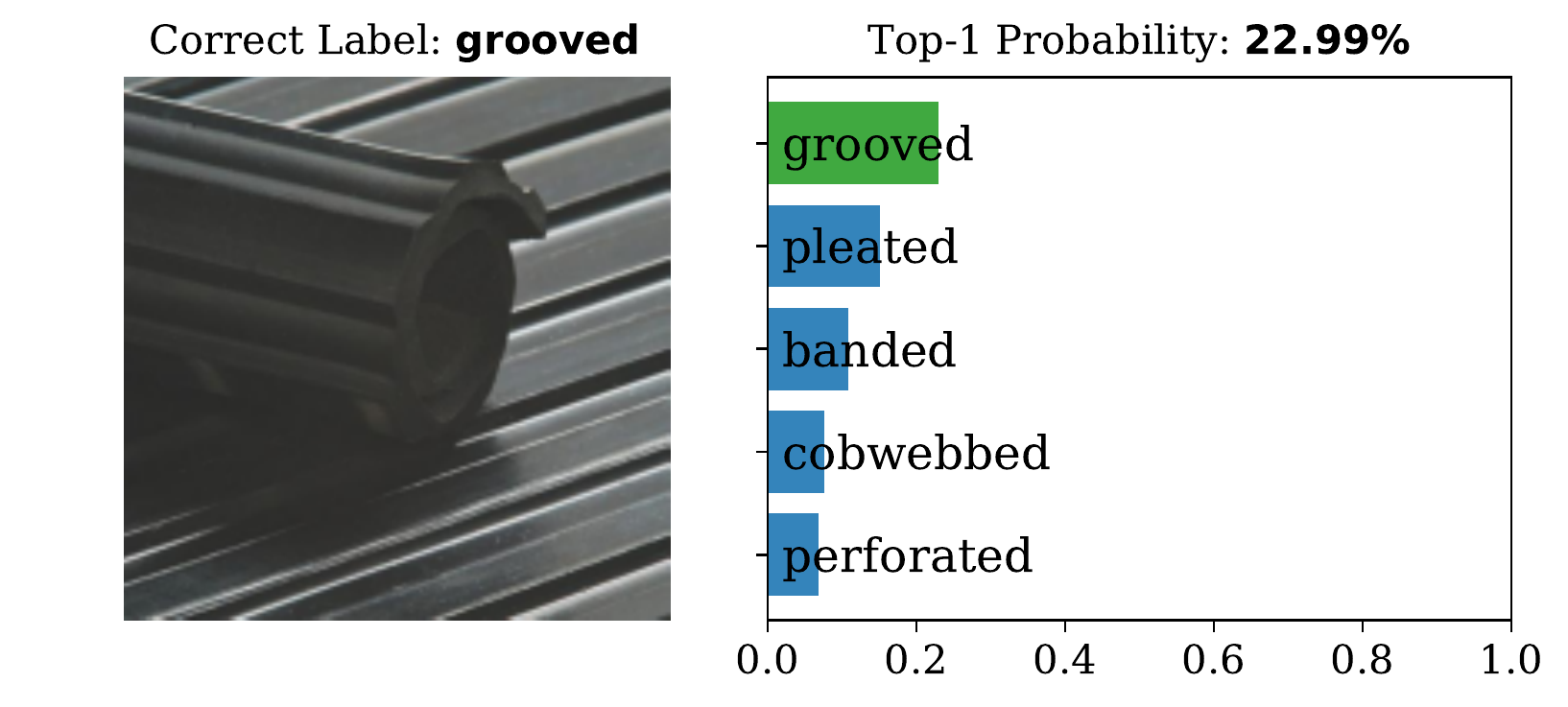}
         \label{fig:c1}
     \end{subfigure}
     \begin{subfigure}[b]{0.48\linewidth}
         \centering
         \includegraphics[width=\textwidth]{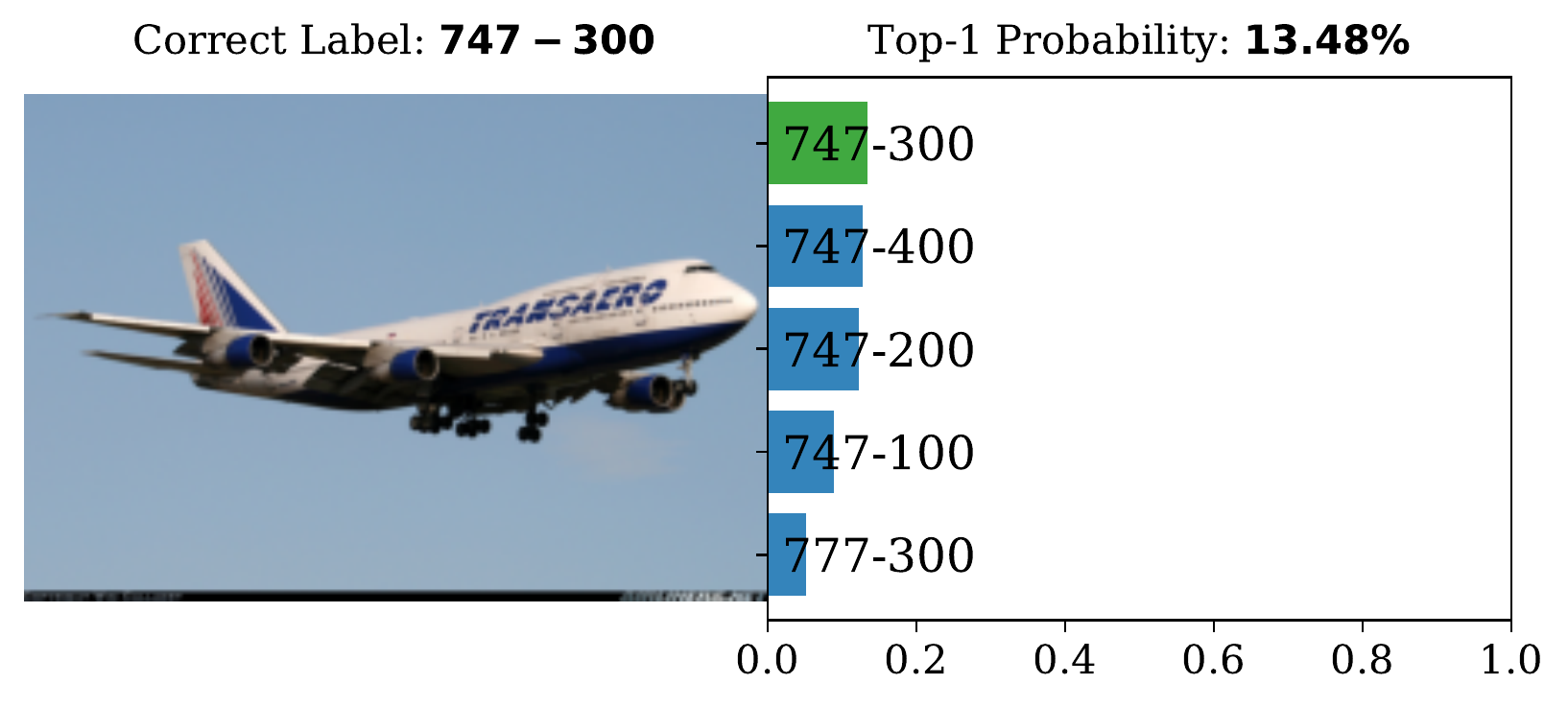}
         \label{fig:w1}
     \end{subfigure}

     \begin{subfigure}[b]{0.48\linewidth}
         \centering
         \includegraphics[width=\textwidth]{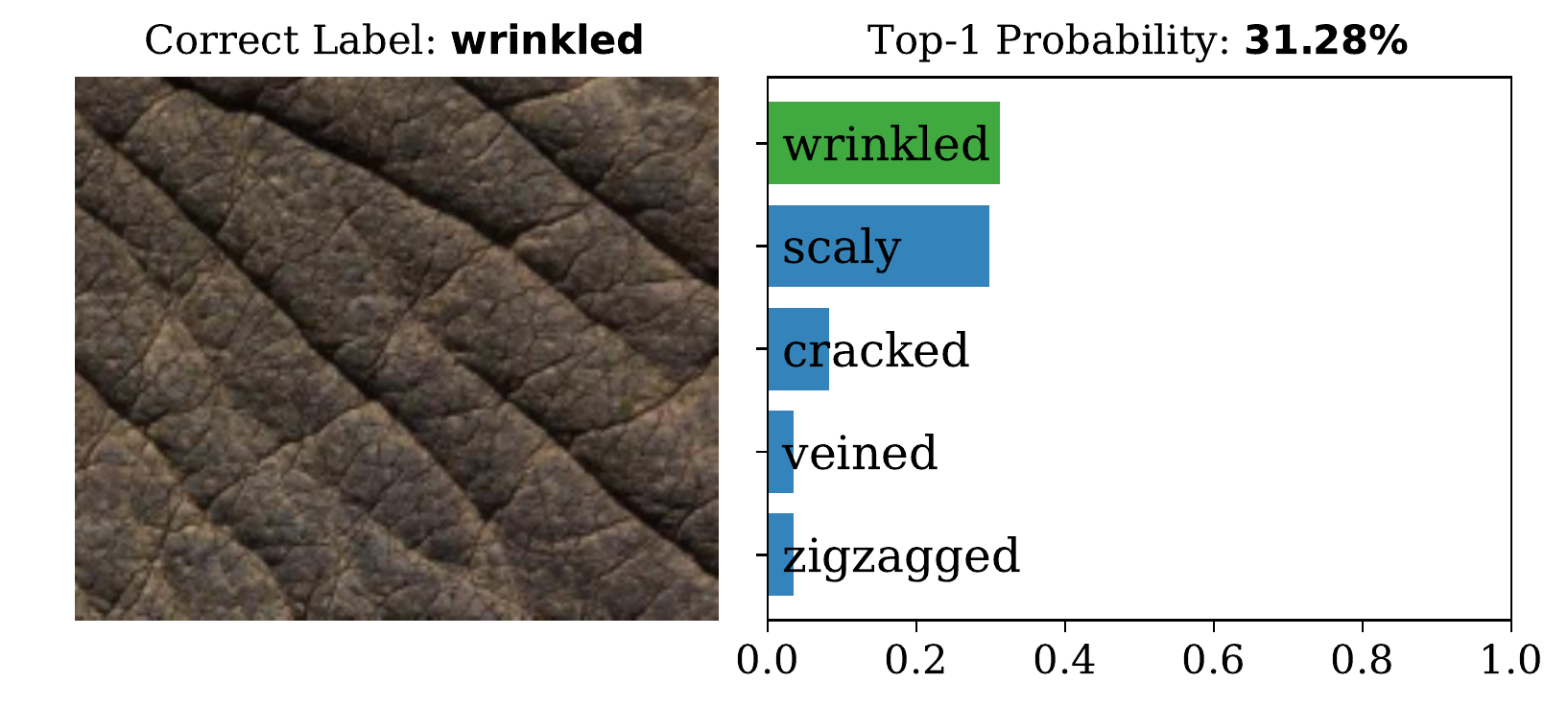}
         \label{fig:c2}
     \end{subfigure}
     \begin{subfigure}[b]{0.48\linewidth}
         \centering
         \includegraphics[width=\textwidth]{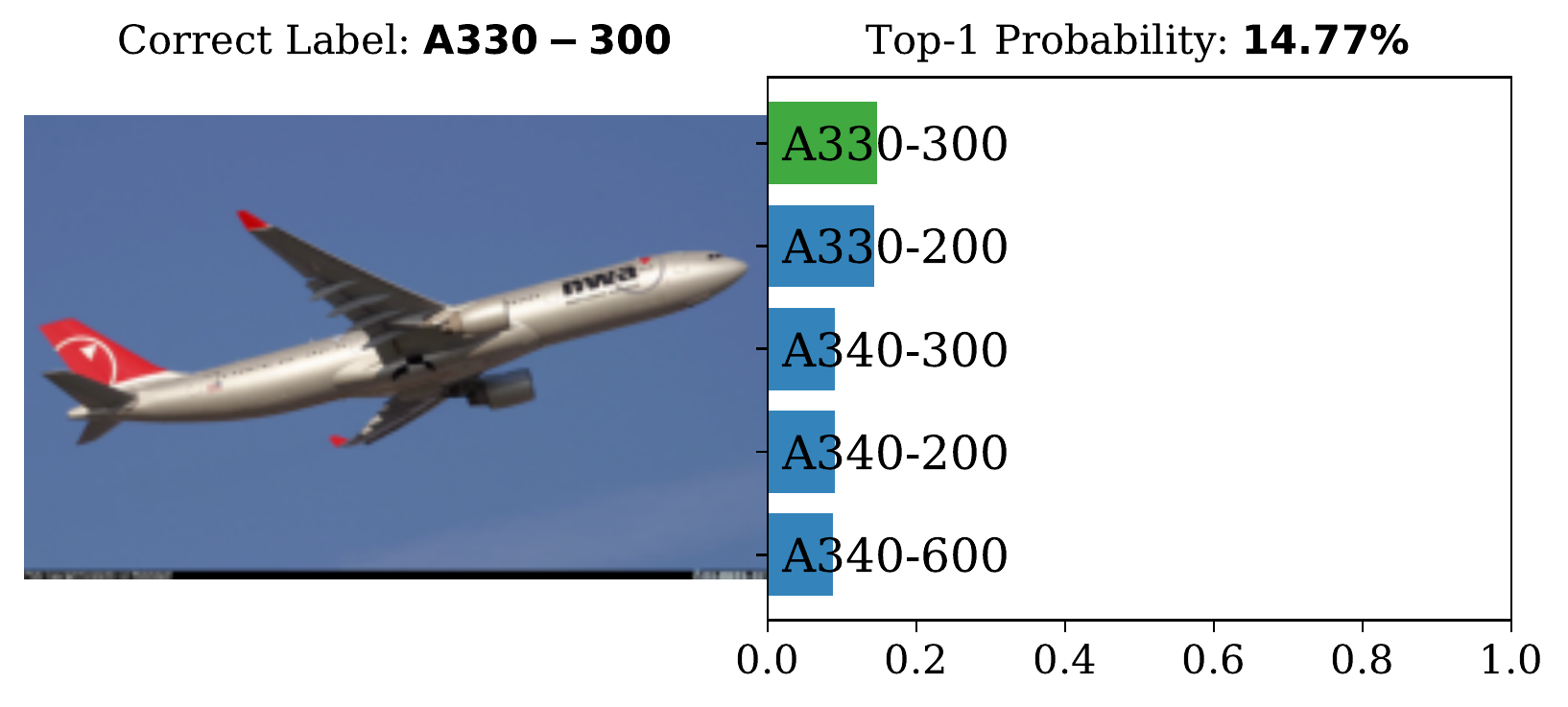}
         \label{fig:w2}
     \end{subfigure}

    \begin{subfigure}[b]{0.48\linewidth}
         \centering
         \includegraphics[width=\textwidth]{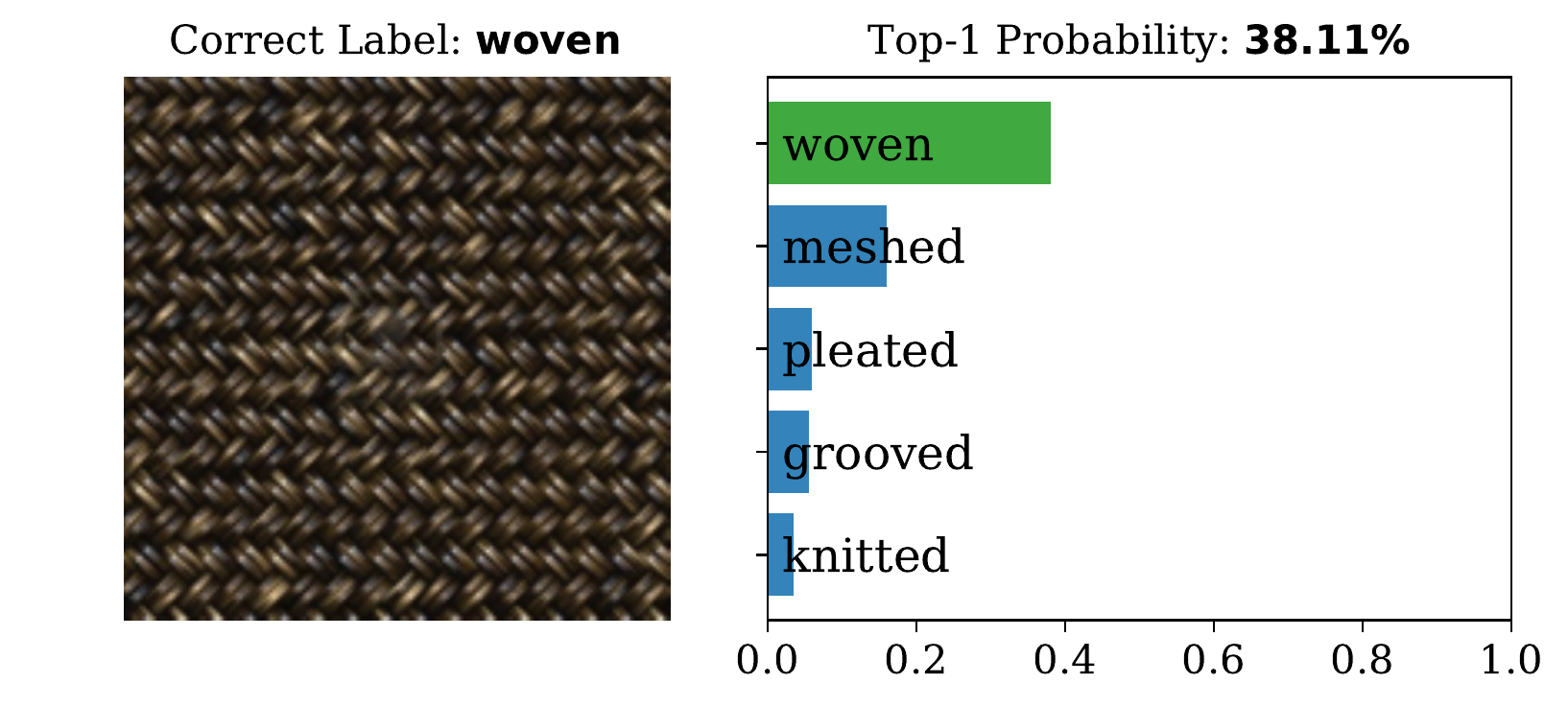}
         \label{fig:c3}
     \end{subfigure}
     \begin{subfigure}[b]{0.48\linewidth}
         \centering
         \includegraphics[width=\textwidth]{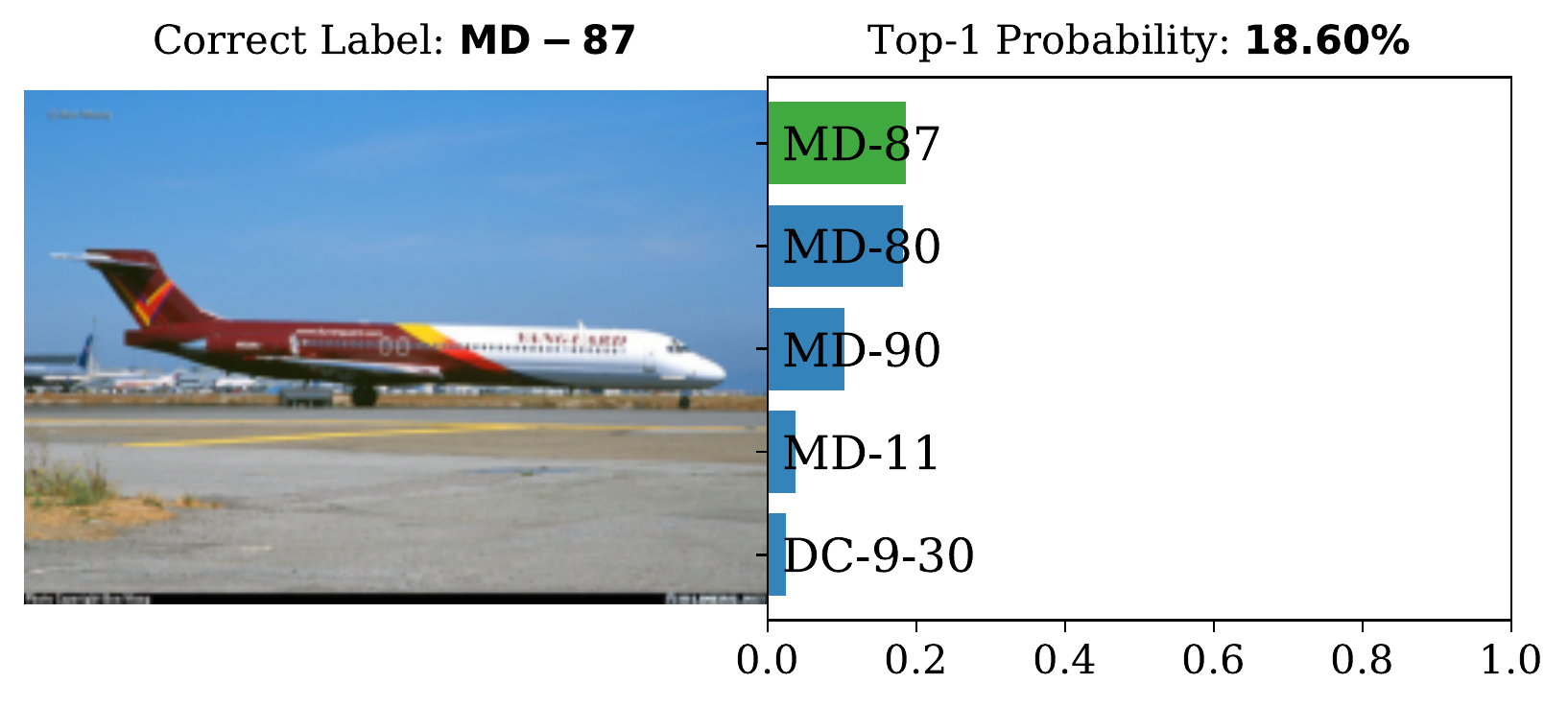}
         \label{fig:w3}
     \end{subfigure}

        \caption{The probability (confidence) can not completely reflect the quality of pseudo labels. Predictions with low confidence may be also correct. We study this phenomenon on DTD (left) and FGVCAircraft (right) datasets.}
        \label{fig:sup_vis}
\end{figure*}

We describe that the probability (confidence) can not completely reflect the quality of pseudo labels the main paper, i.e., unlabeled samples with low confidence scores may be also correctly predicted. Thus we advocate to adopt a top-$K$ pseudo labeling strategy. Here we show some visualization results on DTD and FGVCAircraft datasets in Figure~\ref{fig:sup_vis}. We observe that this phenomenon is more obvious on fine-grained classification datasets, e.g. FGVCAircraft.

\section{Dataset Details and Hand-Craft Prompts}
The details of each dataset and the corresponding hand-craft prompts for pseudo-labeling are provided in Table \ref{datasets_datails}. Note that we only use the simplest prompt to generate pseudo labels without complex prompt engineering.

\begin{table*}[!h]
    \centering
    \begin{tabular}{l|cccc}
    \toprule
        Dataset & Classes & Train & Test & Hand-crafted prompt for pseudo-labeling \\
        \midrule 
        ImageNet & 1,000 & $1.28 \mathrm{M}$ & 50,000 & ``a photo of a [CLASS].'' \\
        Caltech101 & 100 & 4,128 & 2,465 & ``a photo of a [CLASS].'' \\
        OxfordPets & 37 & 2,944 & 3,669 & ``a photo of a [CLASS], a type of pet.'' \\
        StanfordCars & 196 & 6,509 & 8,041 & ``a photo of a [CLASS].'' \\
        Flowers102 & 102 & 4,093 & 2,463 & ``a photo of a [CLASS], a type of flower.'' \\
        Food101 & 101 & 50,500 & 30,300 & ``a photo of [CLASS], a type of food.'' \\
        FGVCAircraft & 100 & 3,334 & 3,333 & ``a photo of a [CLASS], a type of aircraft.'' \\
        SUN397 & 397 & 15,880 & 19,850 & ``a photo of a [CLASS].'' \\
        DTD & 47 & 2,820 & 1,692 & ``[CLASS] texture.'' \\
        EuroSAT & 10 & 13,500 & 8,100 & ``a centered satellite photo of [CLASS].'' \\
        UCF101 & 101 & 7,639 & 3,783 & ``a photo of a person doing [CLASS].'' \\ \bottomrule
    \end{tabular}
    \caption{We show the class number, the size of train/test set and the hand-crafted prompt for pseudo-labeling for each dataset.}
    \label{datasets_datails}
\end{table*}

\end{document}